\documentclass[lettersize,journal]{IEEEtran}
\usepackage{amsmath,amsfonts}
\usepackage{algorithmic}
\usepackage{algorithm}
\usepackage{array}
\usepackage[caption=false,font=normalsize,labelfont=sf,textfont=sf]{subfig}
\usepackage{textcomp}
\usepackage{stfloats}
\usepackage{url}
\usepackage{verbatim}
\usepackage{graphicx}
\usepackage{cite}
\usepackage{enumitem}
\usepackage{multirow}
\usepackage{amsthm,amsmath,amssymb}
\usepackage{mathrsfs}
\usepackage{pifont}
\usepackage{units}
\usepackage{booktabs}
\usepackage[table,xcdraw]{xcolor}
\newtheorem{thm}{Theorem}

\newtheorem{asmp}{Assumption}

\begin{document}
\title{FedBiF: Communication-Efficient Federated Learning via Bits Freezing}
\author{Shiwei Li, Qunwei Li, Haozhao Wang~\IEEEmembership{Member,~IEEE}, Ruixuan Li~\IEEEmembership{Member,~IEEE}, Jianbin Lin, Wenliang Zhong
\thanks{Shiwei Li, Haozhao Wang and Ruixuan Li are with the School of Computer Science and Technology, Huazhong University of Science and Technology, Wuhan 430074, China (e-mail: lishiwei@hust.edu.cn; hz\_wang@hust.edu.cn; rxli@hust.edu.cn). Qunwei Li, Jianbin Lin and Wenliang Zhong are with Ant Group, Hangzhou 310052, China (e-mail: qunwei.qw@antgroup.com; jianbin.ljb@antgroup.com; yice.zwl@antgroup.com).}
\thanks{Corresponding author: Ruixuan Li.}
}



\maketitle

\begin{abstract}
Federated learning (FL) is an emerging distributed machine learning paradigm that enables collaborative model training without sharing local data. Despite its advantages, FL suffers from substantial communication overhead, which can affect training efficiency. Recent efforts have mitigated this issue by quantizing model updates to reduce communication costs. However, most existing methods apply quantization only after local training, introducing quantization errors into the trained parameters and potentially degrading model accuracy.
In this paper, we propose Federated Bit Freezing (FedBiF), a novel FL framework that directly learns quantized model parameters during local training. In each communication round, the server first quantizes the model parameters and transmits them to the clients. FedBiF then allows each client to update only a single bit of the multi-bit parameter representation, freezing the remaining bits. This bit-by-bit update strategy reduces each parameter update to one bit while maintaining high precision in parameter representation.
Extensive experiments are conducted on five widely used datasets under both IID and Non-IID settings. The results demonstrate that FedBiF not only achieves superior communication compression but also promotes sparsity in the resulting models. Notably, FedBiF attains accuracy comparable to FedAvg, even when using only 1 bit-per-parameter (bpp) for uplink and 3 bpp for downlink communication.
The code is available at \url{https://github.com/Leopold1423/fedbif-tpds25}.
\end{abstract}


\section{Introduction}\label{sec:introduction}
\begin{figure*}[t]
\centering
\subfloat[An illustration of FedBiF.]{
    \includegraphics[scale=0.125]{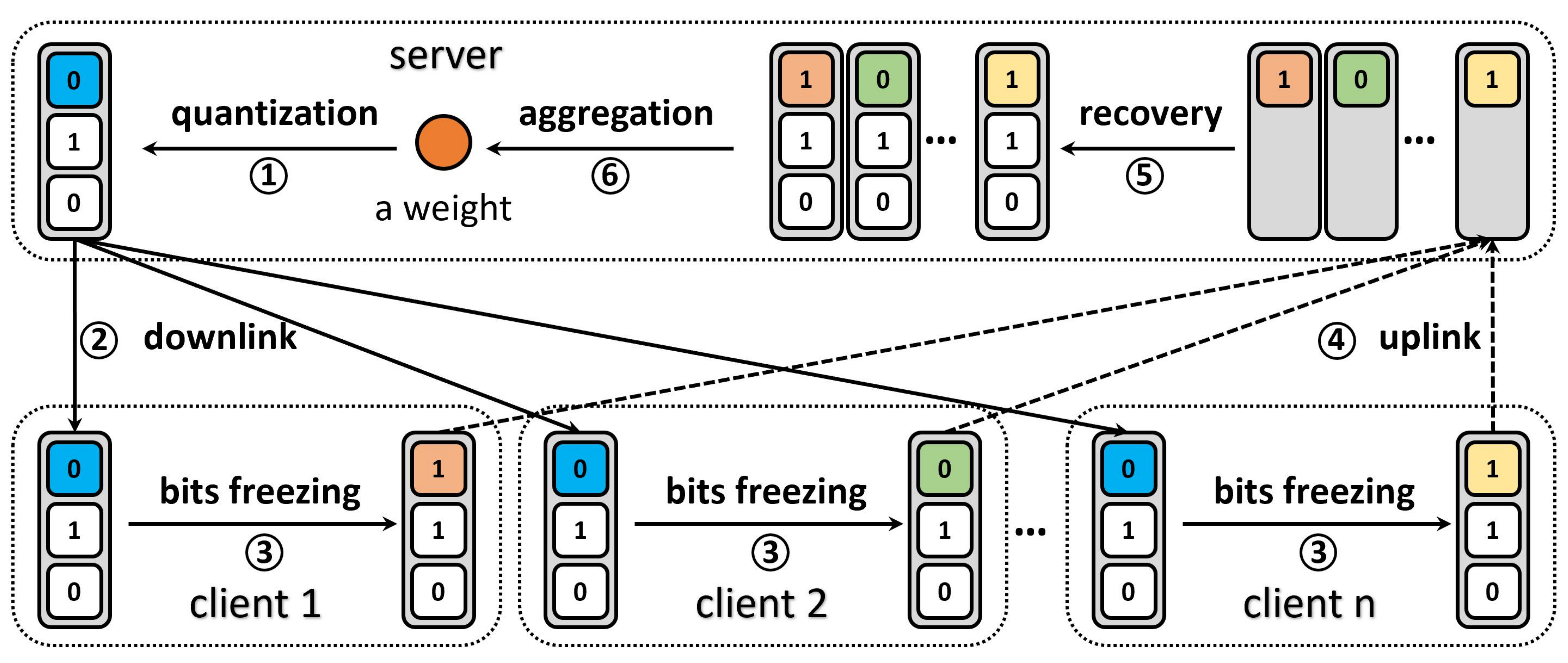}\label{fig:fedbif}
}
\subfloat[An illustration of BiF.]{
    \includegraphics[scale=0.125]{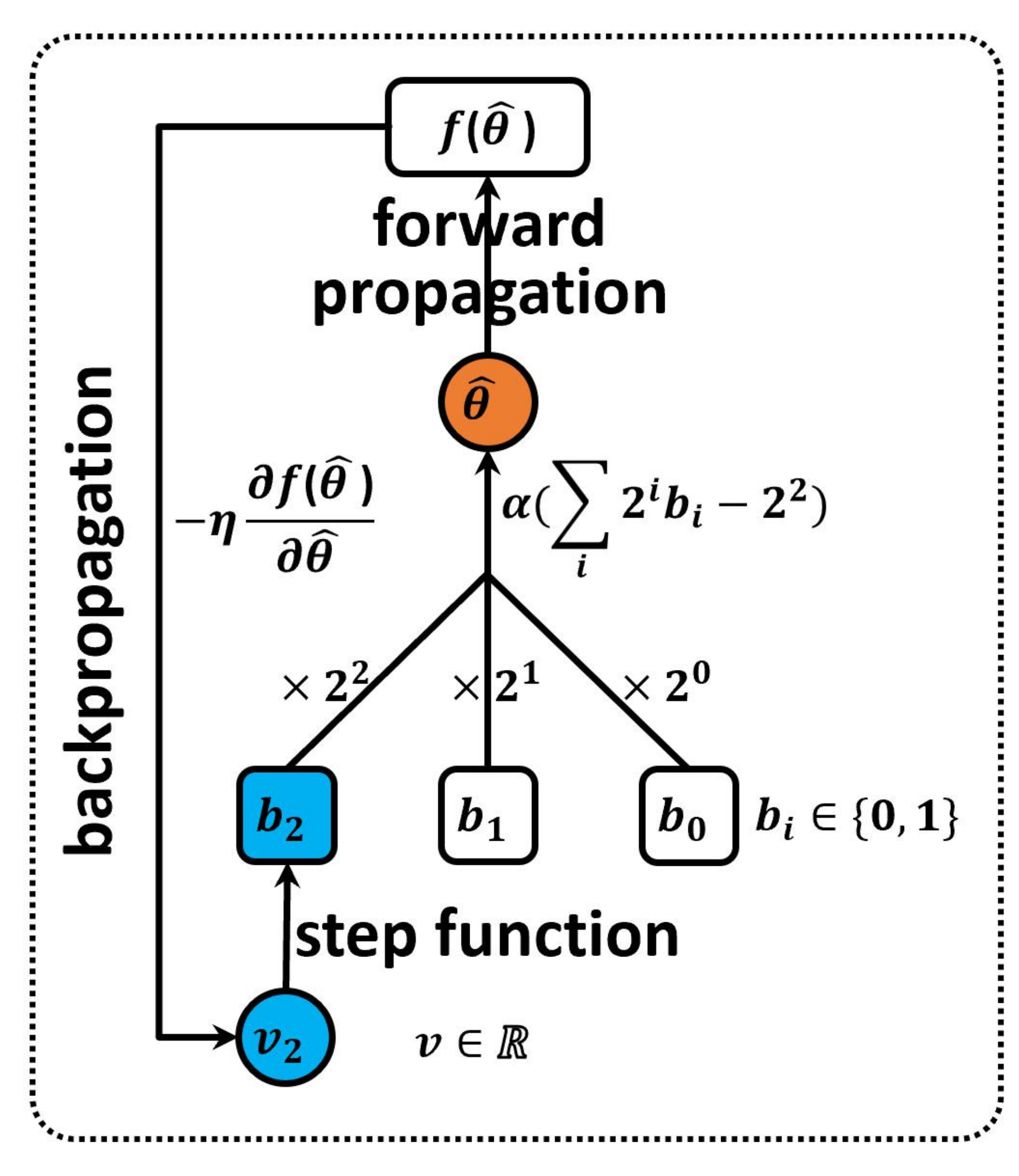}\label{fig:bif}
} 
\caption{(a) The framework of Federated Bit Freezing (FedBiF). \ding{172}: The server quantizes the model parameters; \ding{173}: The server sends the quantized parameters to clients; \ding{174}: Each client trains only the activated bit (depicted in blue) while freezing the remaining bits; \ding{175}: Clients upload the updated activated bits to the server; \ding{176}: The server recovers model parameters with the activated bit and the frozen bits for each client. \ding{177}: The server aggregates the recovered parameters. (b) An illustration of Bit Freezing (BiF). In BiF, each activated bit $b_i$ is learned by optimizing a floating-point virtual bit $v_i$. The step function is defined as $b_i=h(v_i)=1$ if $v_i>0$, and $0$ otherwise. Note that the virtual bit $v_i$ is optimized via gradient descent, with the derivative approximated as $\nicefrac{\partial \hat{\theta}}{\partial v_i}=1$.} 
\end{figure*} 

Driven by the increasing data privacy concerns, federated learning (FL)~\cite{fedavg,DBLP:conf/kdd/WangZH00024,wang2023dafkd} has emerged as a promising machine learning paradigm, which allows distributed clients to collaboratively train a global model without accumulating their data in a centralized repository. The federated training process is generally composed of a series of parameter downloading, local training, parameter uploading, and centralized aggregation phases, which are repeated for several rounds until the model converges~\cite{DBLP:conf/iclr/WangXLX0024}. Each communication round contains one downlink and one uplink communication.

However, the communication latency of iterative model transmissions between the server and the clients can significantly affect the training efficiency, especially when the clients have limited communication bandwidth. Therefore, communication-efficient federated learning (CEFL) has gained great research attention recently, and numerous techniques are adopted to reduce the communication costs~\cite{DBLP:journals/tpds/WangGQLL22,prunefl,fedpara,fedpm,fedpaq}. In this paper, we investigate the application of quantization~\cite{quant-whitepaper} in CEFL.

Quantization aims to represent model parameters using fewer bits, thereby reducing the communication overhead in model synchronization \cite{alpt,lsq}. FedPAQ~\cite{fedpaq} is the pioneer in utilizing quantization on local model updates. Subsequent studies~\cite{adaquantfl,feddq,dadaquant} further improve the compression efficiency by adaptively adjusting the quantization bit width across different training rounds. Alternatively, LFL~\cite{lfl} improves compression by quantizing both local and global model updates simultaneously. 
Despite their achievements in reducing communication costs, a critical limitation remains. Specifically, they employ post-training quantization (PTQ) \cite{quant-whitepaper} on local model updates, where quantization is applied only after local training. This introduces quantization errors into the trained parameters, resulting in information loss for local model updates. In cases where the quantization bit width is extremely small (e.g., 1 bit), the quantization errors can be significant and may lead to a large decrease in model accuracy. 

In this paper, we aim to learn quantized model parameters during local training, thereby eliminating the need for post-training quantization.
Quantization-aware training (QAT) \cite{lsq,quant-whitepaper} is a widely adopted technique for learning quantized parameters. It incorporates a pseudo-quantizer during training, where parameters are quantized before being used in forward propagation.
In FL, QAT is primarily employed to address device heterogeneity \cite{quant_robust} and reduce local computational costs \cite{low-prcison-fl,cocofl,fedaqt} by jointly quantizing both weights and activations, enabling low-precision computation.
However, its effectiveness in reducing communication overhead remains limited. This limitation primarily stems from the inherent trade-off between communication cost and weight precision. For instance, achieving a 1-bit communication cost necessitates quantizing the entire model to 1-bit integers, which significantly compromises the model’s expressiveness.

\begin{figure}
    \centering
    \includegraphics[width=0.99\linewidth]{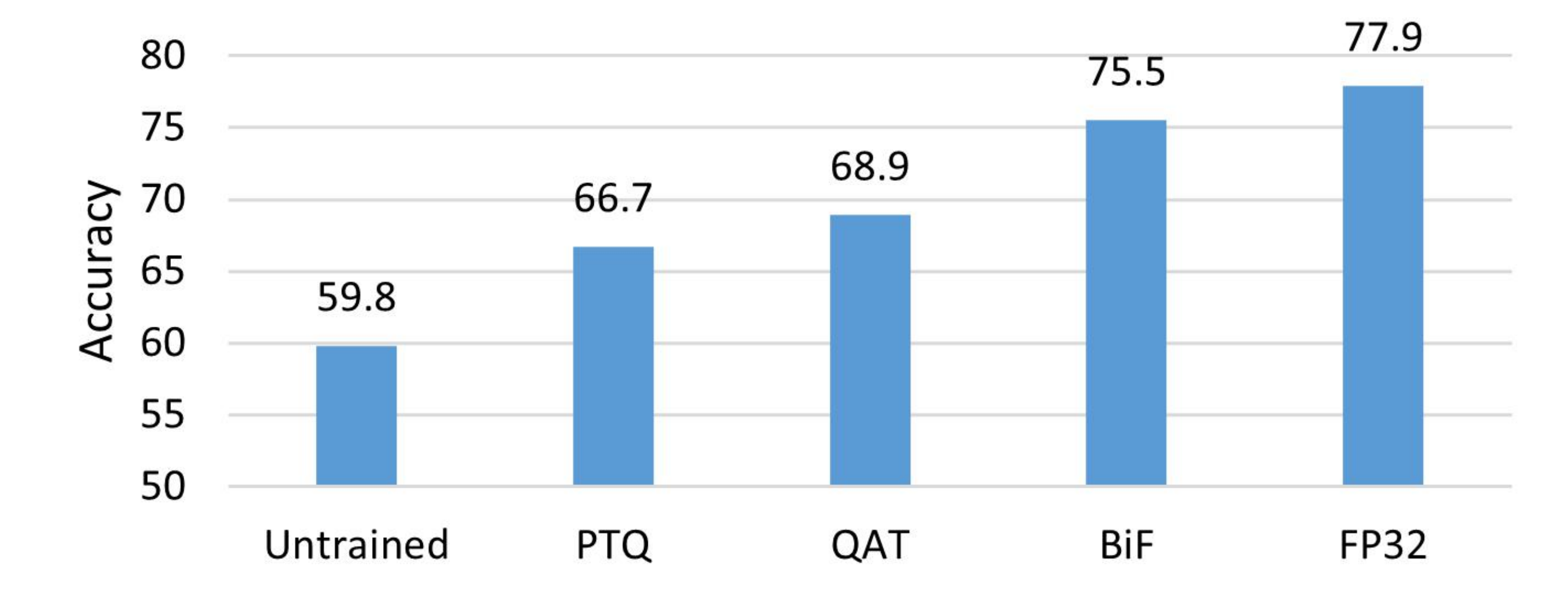}
    \caption{A comparison of Post-Training Quantization (PTQ), Quantization-Aware Training (QAT), and Bits Freezing (BiF). First, a 4-bit quantized model (a 3-layer convolutional neural network) is first initialized on the FMNIST dataset with an initial accuracy of 59.8\% (referred to as the untrained model). The model is then trained for 32 steps under four configurations: FP32, PTQ, QAT, and BiF. 
    FP32 represents full-precision training without any compression. In PTQ, model updates obtained from standard training are quantized to 1 bit. In QAT, a quantizer is integrated into the training process to learn 1-bit parameters. In  BiF, parameters are stored as 4-bit values, but only the second bit is updated during training.
    Results indicate that, under the same communication cost (1 bit-per-parameter), BiF achieves higher accuracy than both PTQ and QAT, and closely approaches the performance of FP32.}\label{fig:motivation}
\end{figure}

To solve this problem, we propose Bits Freezing (BiF). The core idea is that a quantized parameter can be represented using multiple bits. Therefore, optimizing a quantized parameter can be achieved by optimizing these individual bits during local training. Moreover, when certain bits are frozen during training, the transmission of these bits can be omitted, consequently enhancing overall communication efficiency. By doing this, lower bit communication overhead can be achieved while learning high-precision quantized parameters. 
However, the binary nature of these bits, restricted to values of 0 or 1, makes them unsuitable for optimization via gradient-based methods. Therefore, we replace each activated (i.e., to be updated) binary bit with a corresponding floating-point value, referred to as a virtual bit, during local training. 
As shown in Figure \ref{fig:bif}, the binary bit will be obtained by applying the step function on the virtual bits. By optimizing the virtual bit, the binary bit can be effectively learned. 
Figure \ref{fig:motivation} compares PTQ, QAT, and BiF after local training under a 1-bit communication constraint. BiF achieves the highest accuracy within the same communication budget. 
Compared to PTQ, it reduces error by learning the quantization parameters. Compared to QAT, it leverages higher-precision model parameters without introducing additional communication costs.

To this end, we propose \textbf{Federated Bits Freezing (FedBiF)}, a novel framework designed to compress both uplink and downlink communications.
The overall pipeline of FedBiF is illustrated in Figure~\ref{fig:fedbif}.
In each communication round, the server first quantizes the global model and transmits the multi-bit encoded parameters to clients, enabling efficient downlink communication.
FedBiF then allows each client to update only a single bit of the multi-bit parameter representation, freezing the remaining bits. By doing this, the model update can be represented solely by the activated bit which is learned during local training. 
After local training, clients only need to upload the binary values of activated bits to the server where the local models will be recovered and aggregated. 
By iteratively activating different bits in different rounds, FedBiF can learn quantized model parameters effectively while achieving efficient communication. 

Experimentally, we observe that FedBiF achieves greater communication compression than existing methods without compromising model accuracy or convergence speed. 
Moreover, FedBiF is capable of producing sparse models, i.e., certain parameters are naturally optimized to zero. 
Generally, gradient-based algorithms struggle to generate sparse models without manual pruning~\cite{zerofl}. 
In contrast, FedBiF inherently encourages sparsity. Specifically, a parameter is driven to zero when all its associated bits (except for the sign bit) are optimized to zero. Due to the step function, optimizing a bit to zero involves adjusting its corresponding virtual bit to a negative value, respectively.  This is significantly simpler than directly optimizing a parameter to exactly zero. 
It is believed that FedBiF can achieve better model generalization and lead to better accuracy with such model sparsity. 

Our main contributions can be summarized as follows: 
\begin{itemize}[leftmargin=*]
    \item We argue that the post-training manner of existing federated quantization methods may cause accuracy loss when applying relatively small quantization bit widths. In this paper, we propose to learn quantized parameters during local training, representing a pioneering direction for CEFL.
    \item we propose FedBiF, a novel FL framework  in which clients update only a single bit of each multi-bit parameter while freezing the remaining bits. This bit-by-bit update strategy reduces communication to one bit-per-parameter while preserving high precision in parameter representation.
    \item Extensive experiments are conducted on five public datasets. The experimental results reveal not only the superiority of FedBiF in bidirectional communication compression but also the ability of FedBiF to yield sparse models. FedBiF outperforms all baselines, and achieves an accuracy comparable to FedAvg even with 1 bit-per-parameter(bpp) uplink and 3 bpp downlink communication.
\end{itemize}

\section{Related Work}\label{sec:related}
\subsection{Quantization} 
The deep learning community has seen extensive research and applications on quantization techniques, which can be roughly divided into two categories: post-training quantization (PTQ)~\cite{post-aciq,post-adaround}  and quantization-aware training (QAT)~\cite{binaryconnect,lsq,pact}. 
PTQ quantizes a trained model while QAT quantizes the model during training. Specifically, QAT usually quantizes the model parameters in the forward pass and updates the full-precision parameters by the Straight-Through Estimator (STE)~\cite{binaryconnect} during backpropagation. 
In simple terms, STE treats the gradient of operators that are not differentiable or have poor derivative properties as 1, thereby performing effective backpropagation. 
Extensive research has proved that QAT usually enjoys better model accuracy than PTQ~\cite{lsq,quantization}. 
In this paper, we propose a novel quantization training paradigm called Bits Freezing, which trains each bit of the quantized parameters  separately. This method demonstrates significant potential in CEFL, outperforming both PTQ and QAT.

\subsection{Quantization in Federated Learning} 
FedPAQ~\cite{fedpaq} is the pioneer in utilizing quantization on local model updates. 
Recent quantization methods try to enhance their compression ability by adjusting the quantization bit width adaptively during training~\cite{feddq,adaquantfl,dadaquant}.
Considering that early model updates do not need to be particularly precise, 
AdaQuantFL is proposed to employ time-adaptive quantization~\cite{adaquantfl}, using a lower bit width in the initial rounds and gradually increasing the bit width as training proceeds. 
DAdaQuant~\cite{dadaquant} uses a similar time-adaptive quantization and employs client-adaptive quantization, allocating higher bit width to the clients with more data.
On the contrary, FedDQ~\cite{feddq} follows a descending quantization strategy, gradually decreasing the bit width throughout the FL process. 
The rationale behind FedDQ is that the magnitude of model updates gets smaller as the model progressively converges over training, enabling the utilization of lower bit width to effectively represent the local model updates. 
On the other hand, there are also many works that try to quantize both global and local model updates to further improve efficiency.~\cite{docofl,ef21-p,lfl}. 
However, as discussed in Section~\ref{sec:introduction}, the main drawback of these methods lies in the post-training manner. The local model updates are only quantized after local training, and the quantization errors of model parameters inevitably result in a loss of model accuracy. FedBiF overcomes such a drawback by learning quantized parameters during local training, thus requiring no post-training quantization. Recent work, FedBAT \cite{fedbat}, has also introduced a similar mechanism that directly learns the binary model updates during local training.

In addition, while many studies have explored combining QAT with local training in FL, they often achieve only limited communication compression \cite{quant_robust,low-prcison-fl,cocofl,fedaqt}. 
For example, to adapt to the heterogeneous resources, CoCoFL \cite{cocofl} freezes and quantizes several layers, reducing communication and computation, whereas other layers are still trained in full precision.
\cite{low-prcison-fl} proposes an efficient FL paradigm, where the local models are trained with low-precision operations and communicated with the server in low precision format, while only the model aggregation in the server is performed with high-precision computation.
These methods typically reduce local computation and communication overhead by quantizing both activations and model weights. However, the quantization bit width cannot be too low without significantly degrading model accuracy, thus limiting the overall compression effectiveness.
In contrast, FedBiF decouples weight precision from communication overhead. It allows each parameter to retain a multi-bit representation while requiring only a single bit (i.e., the activation bit) to be transmitted during communication.

\subsection{Communication-Efficient Federated Learning} 
Apart from quantization, other techniques have also been employed in CEFL, such as pruning \cite{fedmrn} and low-rank decomposition \cite{li2025fedmud,Wu_2024_MM}. For instance, Adaptive Federated Dropout (AFD) allows clients to train randomly selected subsets of a larger server model’s parameters~\cite{ada-feddropout}. FedPara reduces model size by decomposing weight matrices into smaller components~\cite{fedpara}. Notably, FedMask~\cite{fedmask} and FedPM~\cite{fedpm} achieve 1 bit-per-parameter (bpp) communication by training and transmitting binary masks for each parameter. These methods fix the initial random weights and prune them by learning binary masks. However, they generally fail to match the accuracy of FedAvg~\cite{hidenseek}.
In contrast, the proposed framework, FedBiF, achieves 1 bpp communication by training only one bit within each weight per round. This enables progressive optimization of weights through iterative bit training. Furthermore, FedBiF can automatically prune a weight when all bits, except the sign bit, are optimized to zero.

It is worth noting that this paper primarily focuses on communication compression in the basic federated learning scenario. However, we believe that FedBiF also holds significant potential for application in federated incremental learning \cite{li2025unleashing,li2025fedssi,li2024towards,li2025re} and personalized federated learning \cite{li2024personalized,Wu_2023_ICCV,Wu_2022_IPDPS, Wu_2025_TNNLS}, where efficient communication and adaptive model optimization are equally critical.

\section{Preliminaries}\label{sec:preliminaries}
\subsection{Federated Learning}
Federated learning is built on a client-server topology in which clients are connected to a central server. The general goal of FL is to train a global model by multiple rounds of aggregating locally trained models. 
FedAvg~\cite{fedavg} is a widely used FL algorithm. In the $t$-th round, the server sends the model parameters $\mathbf{w}^{t}$ to selected clients, denoted as $\mathbb{C}^{t}$. 
Each client $C_k$ in $\mathbb{C}^{t}$ performs certain steps of local training on its local datasets $D_k$, and then sends its model update $\mathbf{w}^{t+1}_{k}-\mathbf{w}^{t}$ back to the server. The server aggregates these updates to generate a new global model as follows: 
\begin{equation}\label{eq:fedavg}
    \mathbf{w}^{t+1} = \mathbf{w}^{t} + \sum _{k \in \mathbb{C}^{t}} p_k^t(\mathbf{w}^{t+1}_{k}-\mathbf{w}^{t}), 
\end{equation}
where $p_k^t=\nicefrac{|D_k|}{\sum _{j \in \mathbb{C}^{t}}|D_j|}$ denotes the proportion of the $k$-th client's data to all data participated in the $t$-th round. It is important to note that the primary goal of this paper is to reduce communication overhead. Therefore, the most accurate baseline for comparison is FedAvg without any compression. Our objective is to minimize communication costs while maintaining accuracy close to that of uncompressed FedAvg. For clarity, the notations frequently used throughout this paper are summarized in Table \ref{tab:notation}.

\begin{table}[!htbp]
\centering
\renewcommand\arraystretch{1.1}
\caption{Commonly Used Notations and Descriptions.}
\resizebox{0.48\textwidth}{!}{
\begin{tabular}{cl}
\toprule 
\textbf{Notation} & \textbf{Description} \\ \toprule
$m$         & Bit-width used for quantization. \\
$\alpha$    & Quantization step size. \\
$t$         & Index of the communication round. \\
$D_i$       & Local dataset of the $i$-th client. \\
$\bar{\theta}$ & Quantized integer representation. \\
$\hat{\theta}$ & Quantized floating-point representation. \\
$Q$         & Quantizer. \\
$b_i$       & The $i$-th binary bit. \\
$v_i$       & The $i$-th virtual bit. \\
$s$         & Pre-computed value of frozen bits. \\
$h(\cdot)$  & Step function. \\
$g(\cdot)$  & Mapping from binary bits to floating-point numbers. \\
$f(\cdot)$  & Loss function. \\ 
$\mathbf{w}$ & Model weights. \\
$C$   & Randomly selected clients. \\
$\mathbb{C}$  & Set of randomly selected clients. \\ 
\bottomrule
\end{tabular}
}
\label{tab:notation}
\end{table}

\subsection{Quantization}
Quantization aims to represent full-precision parameters using fewer bits. 
In this paper, we employ the uniform symmetric quantization~\cite{quant-whitepaper} to compress the model parameters. 
Specifically, given the quantization step size $\alpha$ and the quantization bit width $m$, a full-precision tensor $\mathbf{x}$ will be quantized into an $m$-bit integer tensor $\bar{\mathbf{x}}$ as follows:
\begin{equation}\label{eq:quant}
\bar{\mathbf{x}} = Q(\mathbf{x},\alpha,m)=\text{clamp}(\lfloor \frac{\mathbf{x}}{\alpha} \rceil,-2^{m-1}, 2^{m-1}-1), 
\end{equation}
where $\lfloor x \rceil$ rounds $x$ to its nearest integer and ${\text{clamp}}(x,a,b)$ clamps $x$ between $a$ and $b$. The integers $\bar{\mathbf{x}}$ can be de-quantized into the floating-point values $\hat{\mathbf{x}}$ as follows:  
\begin{equation}\label{eq:de-quant}
\hat{\mathbf{x}} = DeQ(\bar{\mathbf{x}},\alpha)=\alpha\cdot\bar{\mathbf{x}}. 
\end{equation}
Notably, we adopt this basic quantization scheme; while more advanced methods could be employed, they are orthogonal to our main contribution. Our focus lies not in the design of the quantizer, but in the bit-freezing mechanism used during the training of quantized parameters.

\section{Methodology}\label{sec:met}
In this section, we present the framework of FedBiF.
As illustrated in Figure \ref{fig:fedbif}, the server transmits the quantized global model to the clients to enable efficient downlink communication. The client then performs bit freezing on the quantized model, training only one activation bit while freezing the remaining ones.
Section~\ref{sec:bif} details the bit freezing process on the client side, and Section~\ref{sec:fedbif} elaborates on the complete FedBiF pipeline. A theoretical analysis of the proposed bit freezing method is provided in Section~\ref{sec:convergence}.

\subsection{Bits Freezing}\label{sec:bif}
During each round, the client receives $m$-bit quantized model parameters $\bar{\mathbf{w}}^t = (\bar{\theta}^{t}, \alpha^{t})$, where $t$ denotes the communication round index.
Here, $\bar{\theta}^{t}$ and $\alpha^{t}$ represent the integer parameters and the quantization step size, respectively. 
Conventionally, these integer parameters $\bar{\theta}^{t}$ are dequantized into floating-point values $\hat{\theta}^{t}$, which are then used to update the client’s local model.
In contrast, our objective is to optimize each individual bit within the integer parameters $\bar{\theta}^{t}$ during local training. To this end, we decompose $\bar{\theta}^{t}$ into a weighted sum of $m$ binary bits $\{b_{m-1}^{t}, \dots, b_{1}^{t}, b_{0}^{t}\}$, as follows:
\begin{equation}\label{eq:bit2int}
\bar{\theta}^{t} = \sum_{i=0}^{m-1}2^{i}\cdot b_{i}^{t}-2^{m-1}.
\end{equation}
The key idea of Bits Freezing (BiF) is to learn the bits $\{b_{m-1}^{t}, \dots, b_{1}^{t}, b_{0}^{t}\}$ individually, thereby determining the optimal integer parameter. However, the binary nature of these bits, restricted to values of 0 or 1, makes them unsuitable for optimization via gradient-based methods. 
Inspired by Binary Neural Networks~\cite{binaryconnect}, we replace each binary bit with a corresponding floating-point value $v_i$, referred to as a virtual bit during local training. The binary values are then obtained as follows:
\begin{equation}\label{eq:step-func}
b_{i} = h(v_{i})= \left\{
\begin{array}{ll}
    1   & \quad \text{if}\quad v_{i}>0,\\ 
    0   & \quad \text{otherwise}.
\end{array}
\right.
\end{equation}

In this way, the virtual bits become the parameters to be optimized during local training. During forward propagation, given the values of all virtual bits, the corresponding estimate of the parameters is computed as:
\begin{equation}\label{eq:vb2fp}
\begin{aligned}
\hat{\theta}^{t}
&= g(\alpha^{t}, v_{0}^{t},...,v_{m-1}^{t})
= \alpha^{t} \cdot \left(\sum_{i=0}^{m-1}2^{i}\cdot h(v_{i}^{t}) - 2^{m-1}\right),
\end{aligned}
\end{equation}
where $\hat{\theta}^{t}$ is used to compute the model output and subsequently derive gradients for optimizing the virtual bits. 
However, as the function $g(\cdot)$ employs a step function and thus has zero gradients almost everywhere, backpropagation from $\hat{\theta}^{t}$ to $v_{i}^{t}$ is not directly feasible.
A common solution is the Straight-Through Estimator (STE)~\cite{ste}, which directly treats $g(\cdot)$ as the identity function during gradient computation, i.e., $\nicefrac{\partial \hat{\theta}^{t}}{\partial v_{i}^{t}} = 1$. Under this approximation, assuming $\nicefrac{\partial f}{\partial \hat{\theta}^{t}}$ denotes the gradient of the objective function $f$ with respect to the floating-point parameters $\hat{\theta}^{t}$, the gradients for the virtual bits become:
\begin{equation}\label{eq:ste_dev}
\frac{\partial f}{\partial v_{i}^{t}} = \frac{\partial f}{\partial \hat{\theta}^{t}} \cdot \frac{\partial \hat{\theta}^{t}}{\partial v_{i}^{t}} = \frac{\partial f}{\partial \hat{\theta}^{t}}.
\end{equation}
It is important to note that updating each activated binary bit necessitates a corresponding floating-point virtual bit. Consequently, training multiple bits simultaneously is inefficient in terms of memory and computation. To address this, we activate only one bit by default and precompute the values of the remaining frozen bits to minimize both memory and computational overhead. 
Specifically, when the $i$-th bit among $m$ bits is activated, the values of all frozen bits can be precomputed and stored as follows:
\begin{equation}\label{eq:s}
s^{t} = \sum_{j\in[m], j\neq i} 2^{j}\cdot b_{j}^{t}-2^{m-1}.
\end{equation}
Subsequently, the forward propagation process described in Eq.(\ref{eq:vb2fp}) can be reformulated as follows:
\begin{equation}
\begin{aligned}
\hat{\theta}^{t}
&= g(\alpha^{t}, v_{i}^{t}, s^t)
= \alpha^{t} \cdot (h(v_{i}^{t}) + s^t),
\end{aligned}
\end{equation}
Compared to conventional training, this method requires storing only a single trainable virtual bit-per-parameter and performing one additional simple addition operation. Moreover, since the frozen bits after addition do not generate gradients, the overall memory consumption remains low.

\subsection{Federated Bits Freezing}\label{sec:fedbif}
In this section, we introduce the proposed bits freezing into FL settings to improve communication efficiency. In FedBiF, the model architecture on the server remains the same as that in typical FL, while local models are designed for training with bits freezing, where a weight parameter is represented by several binary bits. In each round, the server sends the quantized global model to the clients. Given the quantization bit width $m$, the global model parameter $\theta$ is quantized using Eq.~\ref{eq:quant} as follows:
\begin{equation}\label{eq:layer_alpha}
\alpha =\frac{\Vert\theta\Vert_\infty}{2^{m-1}},
\end{equation}
\begin{equation}\label{eq:layer_int}
\bar{\mathbf{\theta}} = Q(\mathbf{\theta}, \alpha, m). 
\end{equation}

Upon receiving the quantized global model, clients can decompose the global integer parameters according to Eq.(\ref{eq:bit2int}) and have the binary bits. Before local training, each client activates a certain bit and initializes its virtual bit with Kaiming initialization~\cite{kaiming-init}. 
Note that the value of each binary bit indicates the sign of its corresponding virtual bit. Therefore, incorporating the signs of the binary bits into the initialization of virtual bits for training would reserve crucial information from the global model. Taking the $i$-{th} bit as an example, and assuming that $b_{i}$ are the binary bits and $v_{i}$ are the initialized virtual bits, then the virtual bits can be adjusted as follows:
\begin{equation}\label{eq:vb_init}
v_{i} = (2\cdot b_{i} -1)\cdot|v_{i}|,
\end{equation}
such that the value of the binary bit would be preserved before the training of virtual bits.

Note that FedBiF activates one bit per round, which not only maximizes communication compression but also helps save storage and computation overhead for local training.
In this paper, we adopt a simple and effective bit selection strategy that cyclically activates each bit of the model parameters during different rounds.  Experimental results will show that randomly activated bits yield comparable performance.
After local training, clients only send the binary values of the activated bits to the server where the local model can be recovered and then aggregated. 
Formally, assuming the bit width is $m$ and the $i$-th bit is activated, the set of selected clients is $\mathbb{C}^{t}$, the aggregation can be formulated as:
\begin{equation}\label{eq:aggregate}
\theta^{t+1} = \alpha^{t}\cdot (2^{i} \sum_{k\in\mathbb{C}^{t}} p_k^t\cdot b_{i,l,k}^{t+1} + s^t),
\end{equation}
where $p_k^t$ denotes the proportion of the $k$-th client's data to all local data used in the $t$-th round, $s^t$ denotes the precomputed sum of all frozen bits as shown in Eq.(\ref{eq:s}).

It is worth noting that an activated bit exhibits one of two behaviors: it either changes or remains the same. If a bit remains unchanged, it indicates that its current value is already accurate, and therefore, no update is required, which is both expected and acceptable. For instance, many studies apply gradient sparsification, where a large number of parameter updates are effectively zero \cite{fetchsgd, zerofl, DBLP:journals/tpds/WangGQLL22, prunefl}. By iterating through all bits during training, if the majority remain unchanged, it can be inferred that the model has converged.

\subsection{Convergence Analysis}\label{sec:convergence}
To analyze the convergence for the proposed method with bits freezing, we consider empirical risk minimization problems as in Eq.(\ref{eq:risk}) following \cite{li2017training}, where $\mathbf{v}\in \mathbb{R}^{m\times d}$ is used to denote all the virtual bits, $m$ is the quantization bit width, and $d$ is the dimension of model parameters $\mathbf{w}$. The loss function is decomposed into a sum of multiple loss functions $\mathbb{F}=\{f_1,f_2,...,f_m\}$:

\begin{equation}\label{eq:risk}
\min_{\mathbf{v}}F(\mathbf{v}) :=\frac{1}{m}\sum _{i=1}^m f_i(\mathbf{v}).
\end{equation}
 At the $t$-th iteration of the gradient descent, we 
select a function $ f^t \in \mathbb{F}$ and update the model parameters as:
\begin{equation}
\mathbf{v}^{t+1} = \mathbf{v}^{t}-\eta^t \nabla  f^t(\mathbf{v}^{t}).
\end{equation}
For the purposes of deriving theoretical guarantees, we make the following assumptions, which are widely used in the optimization community~\cite{li2017training, alpt, convergence}: 
\begin{asmp}\label{asmp:bounded_grads}
The loss function $f_i$ is differentiable and has a bounded gradient, i.e., $\mathbb{E}\Vert\nabla f_i(\mathbf{v})\Vert^2 \leq G^2$.
\end{asmp}
\begin{asmp}\label{asmp:lsmooth}
The loss functions $f_i$ are $L$-smooth, i.e., $\forall \mathbf{v}_1, \mathbf{v}_2 \in \mathbb{R}^{m\times d}$, $\Vert\nabla f_i(\mathbf{v}_1)-\nabla f_i(\mathbf{v}_2)\Vert \leq L\Vert\mathbf{v}_1-\mathbf{v}_2\Vert$ .
\end{asmp}
\begin{asmp}\label{asmp:bounded_bits}
The virtual bits $\mathbf{v}$ have finite diameter, i.e., $\forall \mathbf{v}_1, \mathbf{v}_2 \in \mathbb{R}^{m\times d}$, $\Vert\mathbf{v}_1 - \mathbf{v}_2\Vert\leq D$.
\end{asmp}
\begin{asmp}\label{asmp:bounded_weights}
The model parameters $\mathbf{w}$ have a bounded norm, i.e.,  $\Vert\mathbf{w}\Vert_\infty\leq P$.
\end{asmp}
\begin{thm}[Convex]\label{thm:1}
Assume the learning rate decays as $\eta ^t=\frac{c }{\sqrt{t}}$, for a constant $c$. At the $T$-th iteration, for bits freezing with all $m$ bits activated and with convex loss functions, we have: 
\begin{equation}
\begin{aligned}
    \mathbb{E}\left[F(\bar{\mathbf{v}}^T)-F(\mathbf{v}^*)\right] \leq 
    & {\frac{D^2P+2c^2G^2P}{c\sqrt{T}2^{m+2}}} + \frac{\sqrt{md}LDP^2}{2^{2m+1}},
    \end{aligned}
\end{equation}
\end{thm}
\noindent where $\bar{\mathbf{v}}^T=\frac{1}{T}\sum_{t=1}^T \mathbf{v}^t$ and $\mathbf{v}^*={\arg\min}_{\mathbf{v}} F(\mathbf{v})$.

\begin{thm}[Non-convex]\label{thm:2}
Assume the learning rate is set to $\eta ^t=\frac{c }{\sqrt{T}}$, for a constant $c$, where $T$. At the $T$-th iteration, for bits freezing with all $m$ bits activated and with non-convex loss functions, we have: 
\begin{equation}
\begin{aligned}
    \frac{1}{T} \sum_{t=0}^{T-1} \mathbb{E}\left\Vert F(\mathbf{v}^t)\right\Vert ^2 
    & \leq \frac{ 2P( F(\mathbf{v}^{0}) - F(\mathbf{v}^{*})) + PLG^2c^2}{2^{m+2}c\sqrt{T}}  \\ & \frac{P^2GL\sqrt{md}}{2^{2m+1}} \\
    \end{aligned}
\end{equation}
\end{thm}

Theorems \ref{thm:1} and \ref{thm:2} demonstrate that a model trained using bits freezing exhibits convergence up to an error floor, which correlates with the quantization bit width $m$. With the increase in the bit width, the error floor experiences exponential decay. The proof is provided in the Appendix.

\section{Experiments}\label{sec:exp}
\subsection{Experimental Setup}

\textbf{Datasets and Models.} 
We evaluate FedBiF on five widely used image classification datasets: FMNIST (Fashion-MNIST)\cite{fmnist}, SVHN\cite{svhn}, CIFAR-10~\cite{cifar10}, CIFAR-100~\cite{cifar10}, and TinyImageNet~\cite{tinyimagenet}. To assess performance across different architectures, we use a CNN with four convolutional layers and one fully connected layer for FMNIST and SVHN, and ResNet-18~\cite{resnet} for CIFAR-10, CIFAR-100, and TinyImageNet. Additionally, we validate the performance of our method on CIFAR-10 using DenseNet40~\cite{densenet}. Batch normalization (BN)\cite{bn} is applied to stabilize training, and ReLU\cite{relu} is used as the activation function.

\textbf{Data Partitioning.} 
We consider both scenarios of IID and Non-IID data distribution, referring to the data partitioning benchmark of FL~\cite{noniid-benchmark}. Under IID partitioning, an equal quantity of data is randomly sampled for each client. The Non-IID scenario further encompasses two distinct label distributions for partitioning, termed Non-IID-1 and Non-IID-2. In Non-IID-1, the proportion of the same label among clients follows the Dirichlet distribution~\cite{dirichlet}. 
In contrast, Non-IID-2 involves clients with datasets containing only a subset of labels. Specifically, for Non-IID-1, we fix the Dirichlet parameter at 0.3, while for Non-IID-2, each client contains only randomly chosen 30\% of the labels.

\textbf{Baselines.} 
FedAvg is adopted as the backbone algorithm in our experiments. 
Note that the primary goal of this paper is to reduce communication overhead. Therefore, the most accurate baseline for comparison is FedAvg, which does not perform any compression.
We further compare FedBiF with several state-of-the-art quantization methods, including SignSGD~\cite{signsgd}, FedPAQ~\cite{fedpaq}, FedDQ~\cite{feddq}, DAdaQuant~\cite{dadaquant} and LFL~\cite{lfl}. 
SignSGD only transfers the signs of the local model updates to the server. Specifically, a model update $\Delta \mathbf{w}$ is binarized into $\Delta \hat{\mathbf{w}}=\alpha\cdot\text{sign}(\Delta x^{t+1}_{k})$. FedPAQ quantizes the model updates by a given quantization bit width, while FedDQ quantizes the model updates with a given quantization step size. DAdaQuant dynamically adjusts the bit width used by each client in each round by monitoring the local training loss and the local dataset size. It consists of two parts: time-adaptive quantization and client-adaptive quantization. Let us first denote the set of selected clients in the $t$-th round as $\mathbb{C}^{t}$, the proportion of the $k$-th client's data quantity to the total data quantity as $p_k$, and the training loss of the $k$-th client as $l_k$. 
In time-adaptive quantization, server tracks a running average loss $F^{t}=\psi F^{t-1}+(1-\psi)G^{t}$, where $G^{t}=\sum_{k\in\mathbb{C}^{t}}p_k\cdot l_{k}^{t}$. 
The server determines training convergence whenever $F^{t}\ge F^{t+1-\phi}$. On convergence, the quantization bit width is doubled and then fixed for at least $\phi$ rounds.
In client-adaptive quantization, assuming the bit width of time-adaptive quantization is $m$, the bit width of the $k$-th client is set to $m_k = p_k^{2/3} \cdot \sqrt{\sum_{k\in\mathbb{C}^{t}}p_{k}^{2/3}/\sum_{k\in\mathbb{C}^{t}}p_{k}^{2}/m^{2}}$. LFL performs stochastic quantization proposed in QSGD \cite{qsgd} on both local and global model updates.

\textbf{Hyperparameters.} 
In our experiments, we set the number of clients to 100, with 10 clients randomly selected to participate in each round of training. The open-sourced FL framework Flower~\cite{flower} is employed to simulate FL within a single machine. The batch size is 64, and the number of local epochs is 3. SGD~\cite{sgd} is employed as the local optimizer, and the learning rate is tuned among \{1.0, 0.1, 0.01, 0.001\} for the best outcomes. For FMNIST and SVHN, we set the learning rate to 0.01 and train the model for 100 rounds. For CIFAR-10, CIFAR-100, and TinyImageNet, we set the learning rate to 0.01 and train the model for 200 rounds. For communication compression methods, we tune the hyperparameters to achieve maximal compression while preserving model accuracy.
For FedBiF, we tune the quantization bit width of downlink communication among \{2, 3, 4\}. 
For SignSGD, we tune the step size $\alpha$ among \{0.01, 0.001, 0.0001\} and finally set it to 0.001. 
For FedPAQ, we tune the bit width $m$ among \{2, 3, 4\} and finally set it to 4. For FedDQ, we tune the quantization step size $\alpha$ among \{0.01, 0.003, 0.001\} and finally set it to 0.003. 
Referring to DAdaQuant~\cite{dadaquant}, we set $\psi$ to 0.9, $\phi$ to one tenth of the number of total training rounds. Also, the quantization bit width of DAdaQuant adjusts itself during training, which starts at 1, and we limit it such that it does not exceed the quantization bit width used by FedPAQ (i.e., 4). For LFL, we set its quantization bit width to achieve the same communication overhead as FedBiF.
Each experiment is run five times on Nvidia 3090 GPUs with Intel Xeon E5-2673 CPUs. The average results and standard deviation are reported. 

\subsection{Overall Performance}\label{sec:performance}
\begin{table*}[ht]
\centering
\caption{The test accuracy and communication costs of various methods on five datasets.}\label{tb:performance}
\label{tab:glue_results}
\renewcommand{\arraystretch}{1.0}
    
\resizebox{0.9\textwidth}{!}{
\begin{tabular}{llcccccc}
\toprule[1pt]
& & \multicolumn{4}{c}{Test Accuracy} & \multicolumn{2}{c}{\# Bits Per Parameter} \\ 
\cmidrule(r){3-6} \cmidrule(r){7-8}  

& 
& \makebox[0.01\textwidth][c]{IID} & \makebox[0.01\textwidth][c]{Non-IID-1} 
& \makebox[0.01\textwidth][c]{Non-IID-2} & \makebox[0.08\textwidth][c]{Accuracy Loss} & \makebox[0.05\textwidth][c]{Uplink} & \makebox[0.05\textwidth][c]{Downlink}  
\\ 
\toprule\multirow{7}{*}{FMNIST (CNN)} 
 & FedAvg    & 88.6 ($\pm$0.2) & 86.6 ($\pm$0.3) & 82.6 ($\pm$0.2) & +0.00 & 32 & 32 \\
 & SignSGD   & 86.4 ($\pm$0.2) & 84.1 ($\pm$0.1) & 80.8 ($\pm$0.6) & -2.18 & 1.0  & 32 \\
 & FedPAQ    & 88.5 ($\pm$0.1) & 86.3 ($\pm$0.1) & 81.9 ($\pm$0.3) & -0.38 & 4.0  & 32 \\
 & FedDQ     & 88.2 ($\pm$0.1) & 85.6 ($\pm$0.2) & 82.3 ($\pm$0.3) & -0.59 & 2.4  & 32 \\
 & DAdaQuant & 88.4 ($\pm$0.2) & 86.0 ($\pm$0.2) & 82.2 ($\pm$0.3) & -0.41 & 2.7  & 32 \\
 & LFL       & 85.9 ($\pm$0.1) & 83.1 ($\pm$0.2) & 81.6 ($\pm$0.6) & -2.41 & 1.0  & 3.0  \\
 & FedBiF    & 89.0 ($\pm$0.1) & 86.3 ($\pm$0.1) & 82.3 ($\pm$0.4) & -0.07 & 1.0  & 3.0  \\
\toprule\multirow{7}{*}{SVHN (CNN)} 
 & FedAvg    & 88.8 ($\pm$0.2) & 85.1 ($\pm$0.5) & 84.0 ($\pm$0.2) & +0.00 & 32 & 32 \\
 & SignSGD   & 86.8 ($\pm$0.1) & 81.8 ($\pm$0.3) & 80.8 ($\pm$0.7) & -2.87 & 1.0  & 32 \\
 & FedPAQ    & 87.0 ($\pm$0.2) & 82.7 ($\pm$0.1) & 80.3 ($\pm$0.7) & -2.66 & 4.0  & 32 \\
 & FedDQ     & 86.6 ($\pm$0.3) & 82.5 ($\pm$0.7) & 79.2 ($\pm$0.8) & -3.20 & 2.8  & 32 \\
 & DAdaQuant & 86.1 ($\pm$0.3) & 81.7 ($\pm$0.3) & 80.7 ($\pm$0.3) & -3.14 & 2.5  & 32 \\
 & LFL       & 86.5 ($\pm$0.2) & 83.1 ($\pm$0.4) & 79.7 ($\pm$0.6) & -2.90 & 1.0  & 3.0  \\
 & FedBiF    & 88.9 ($\pm$0.1) & 85.2 ($\pm$0.2) & 83.9 ($\pm$0.6) & +0.01 & 1.0  & 3.0  \\
\toprule\multirow{7}{*}{CIFAR-10 (ResNet18)} 
 & FedAvg    & 78.1 ($\pm$0.2) & 70.1 ($\pm$0.4) & 64.5 ($\pm$0.1) & +0.00 & 32 & 32 \\
 & SignSGD   & 67.5 ($\pm$0.5) & 60.6 ($\pm$0.8) & 55.2 ($\pm$0.1) & -9.78 & 1.0  & 32 \\
 & FedPAQ    & 73.0 ($\pm$0.8) & 64.9 ($\pm$0.3) & 59.8 ($\pm$0.1) & -4.95 & 4.0  & 32 \\
 & FedDQ     & 70.7 ($\pm$0.1) & 64.5 ($\pm$0.6) & 58.9 ($\pm$0.7) & -6.18 & 2.7  & 32 \\
 & DAdaQuant & 71.3 ($\pm$0.1) & 64.8 ($\pm$0.1) & 58.8 ($\pm$1.1) & -5.91 & 2.6  & 32 \\
 & LFL       & 70.4 ($\pm$0.4) & 64.7 ($\pm$0.7) & 60.4 ($\pm$0.2) & -5.74 & 1.0  & 4.0  \\
 & FedBiF    & 77.4 ($\pm$0.2) & 69.9 ($\pm$0.7) & 64.8 ($\pm$0.3) & -0.17 & 1.0  & 4.0  \\
\toprule\multirow{7}{*}{CIFAR-100 (ResNet18)} 
 & FedAvg    & 40.7 ($\pm$0.2) & 38.3 ($\pm$0.3) & 39.0 ($\pm$0.1) & +0.00 & 32 & 32 \\
 & SignSGD   & 34.5 ($\pm$1.1) & 33.3 ($\pm$0.3) & 32.2 ($\pm$0.3) & -6.00 & 1.0  & 32 \\
 & FedPAQ    & 39.8 ($\pm$0.3) & 36.9 ($\pm$0.1) & 36.7 ($\pm$0.1) & -1.53 & 4.0  & 32 \\
 & FedDQ     & 35.6 ($\pm$0.2) & 34.7 ($\pm$0.3) & 33.4 ($\pm$0.3) & -4.74 & 3.1  & 32 \\
 & DAdaQuant & 38.4 ($\pm$0.3) & 36.7 ($\pm$0.2) & 36.4 ($\pm$0.3) & -2.17 & 2.2  & 32 \\
 & LFL       & 36.0 ($\pm$0.3) & 35.6 ($\pm$0.3) & 34.3 ($\pm$0.7) & -4.03 & 1.0  & 4.0  \\
 & FedBiF    & 41.3 ($\pm$0.1) & 38.7 ($\pm$0.2) & 39.1 ($\pm$0.4) & +0.38 & 1.0  & 4.0  \\
\toprule\multirow{7}{*}{TinyImageNet (ResNet18)} 
 & FedAvg    & 38.3 ($\pm$0.1) & 36.5 ($\pm$0.9) & 34.7 ($\pm$0.6) & +0.00 & 32 & 32 \\
 & SignSGD   & 32.2 ($\pm$0.1) & 31.8 ($\pm$0.3) & 29.8 ($\pm$0.3) & -5.21 & 1.0  & 32 \\
 & FedPAQ    & 35.1 ($\pm$0.8) & 34.4 ($\pm$0.6) & 32.2 ($\pm$0.2) & -2.60 & 4.0  & 32 \\
 & FedDQ     & 33.1 ($\pm$0.7) & 32.8 ($\pm$0.3) & 30.7 ($\pm$0.3) & -4.31 & 3.3  & 32 \\
 & DAdaQuant & 33.8 ($\pm$0.1) & 32 ($\pm$0.5) & 29.9 ($\pm$0.8) & -4.61 & 2.3  & 32 \\
 & LFL       & 33.5 ($\pm$0.3) & 33.7 ($\pm$0.3) & 31.4 ($\pm$0.1) & -3.63 & 1.0  & 4.0  \\
 & FedBiF    & 38.5 ($\pm$0.3) & 36.2 ($\pm$0.3) & 34.7 ($\pm$0.1) & -0.04 & 1.0  & 4.0  \\
\bottomrule[1pt]
\end{tabular}
}
\parbox{0.9\textwidth}{\vspace{5pt}
    Since the number of parameters to transmit remains constant for different methods, we employ the bit-per-parameter (bpp) as the standard unit for measuring communication costs. The average bpp for uplink and downlink communication are shown in the corresponding columns.
}
\end{table*}

\begin{figure*}
\centering
\subfloat[CIFAR-10, IID]{
    \includegraphics[scale=0.29]{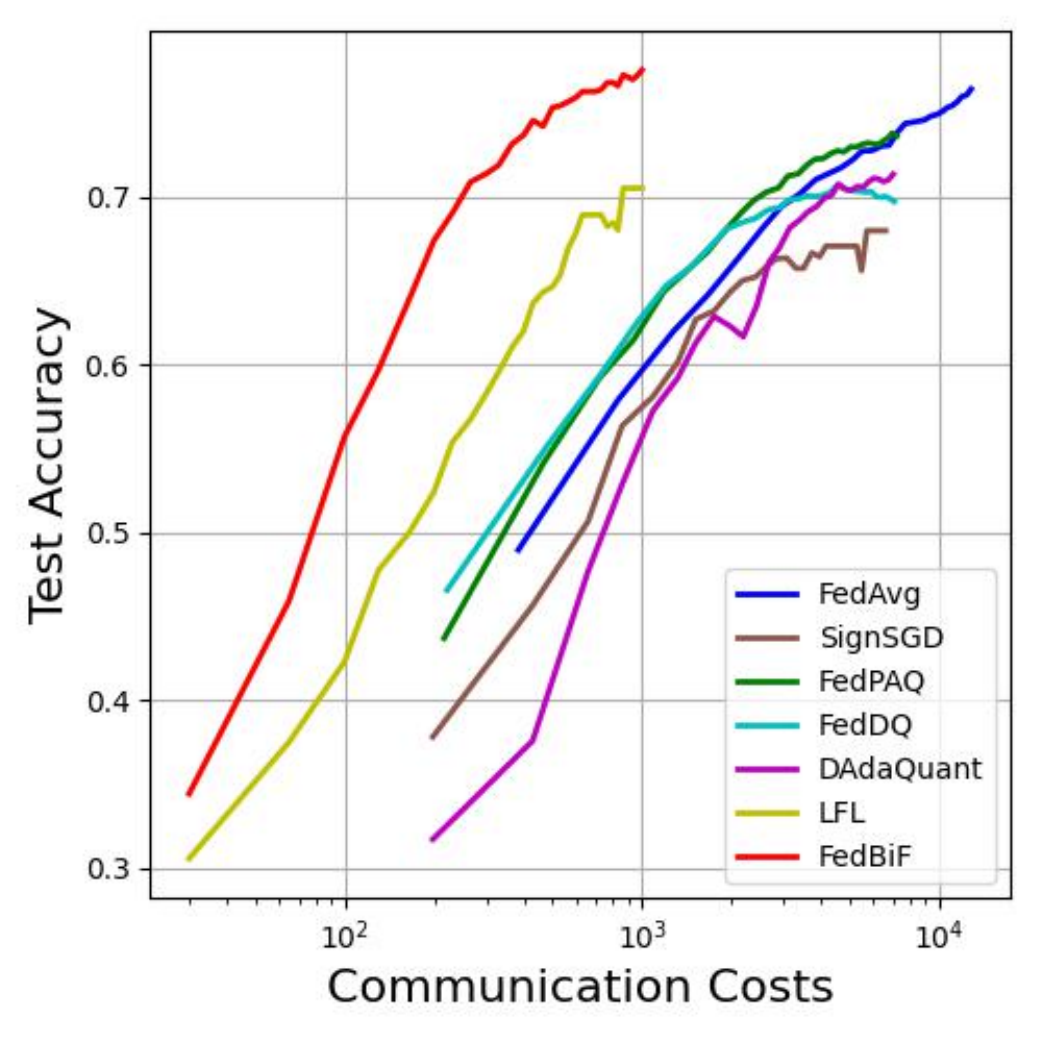}
}
\subfloat[CIFAR-10, Non-IID-1]{
    \includegraphics[scale=0.29]{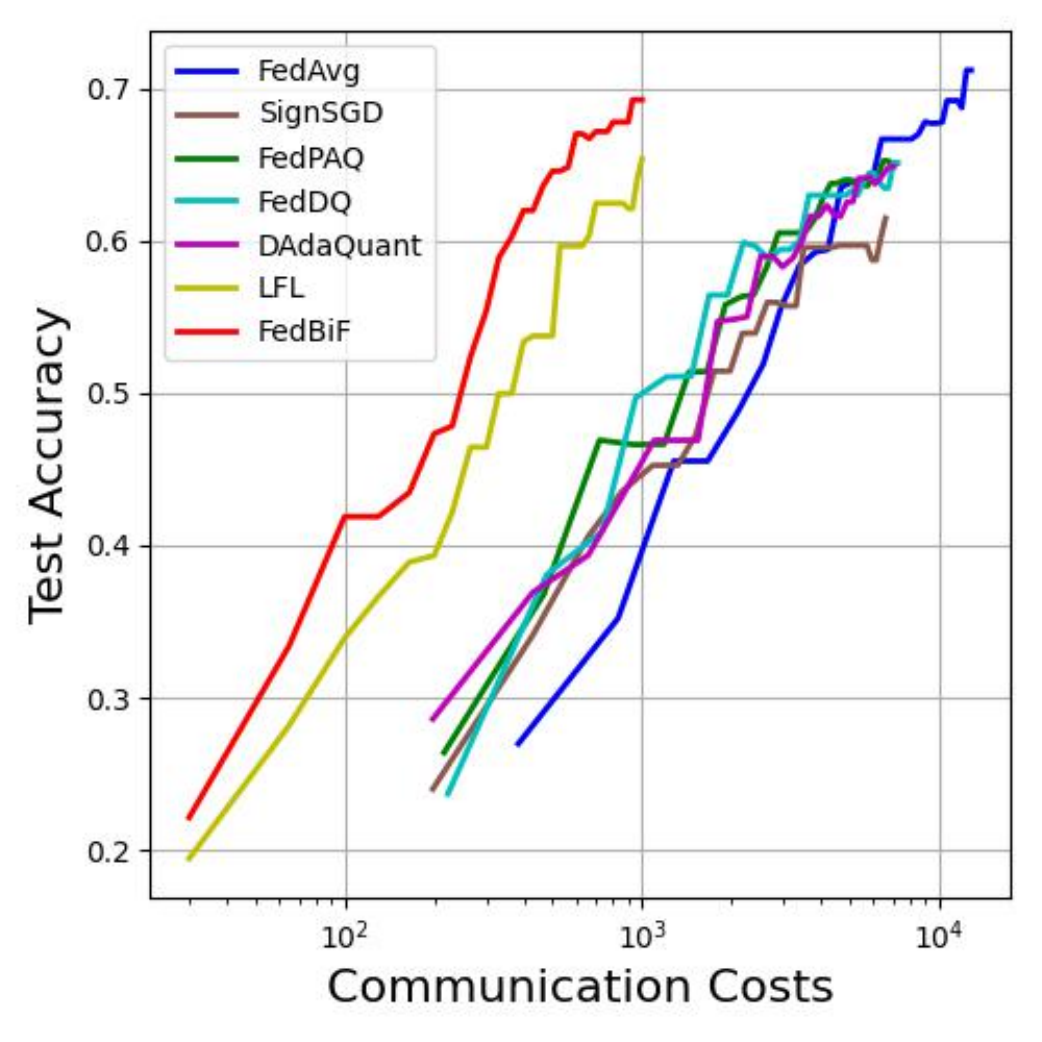}
}
\subfloat[CIFAR-10, Non-IID-2]{
    \includegraphics[scale=0.29]{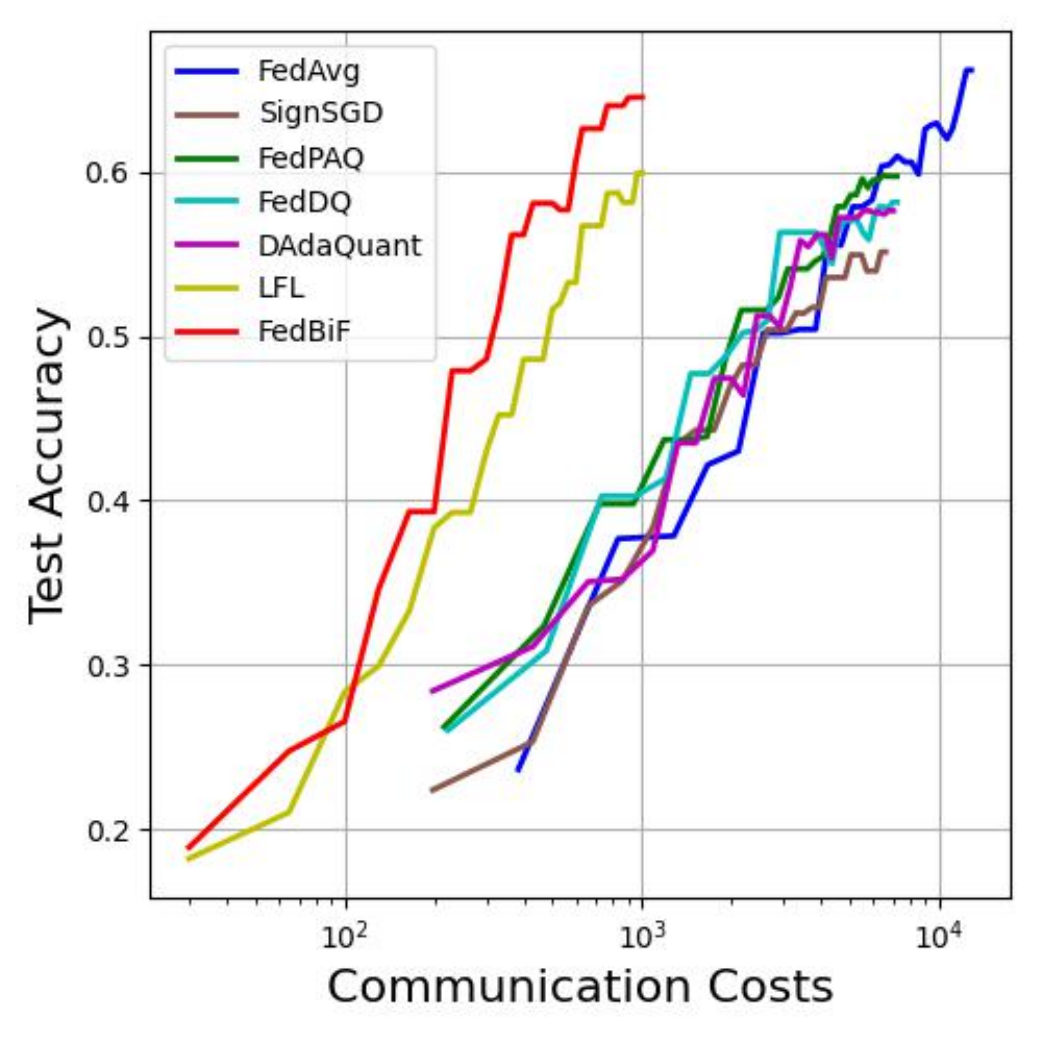}
}
\hfill
\subfloat[CIFAR-100, IID]{
    \includegraphics[scale=0.29]{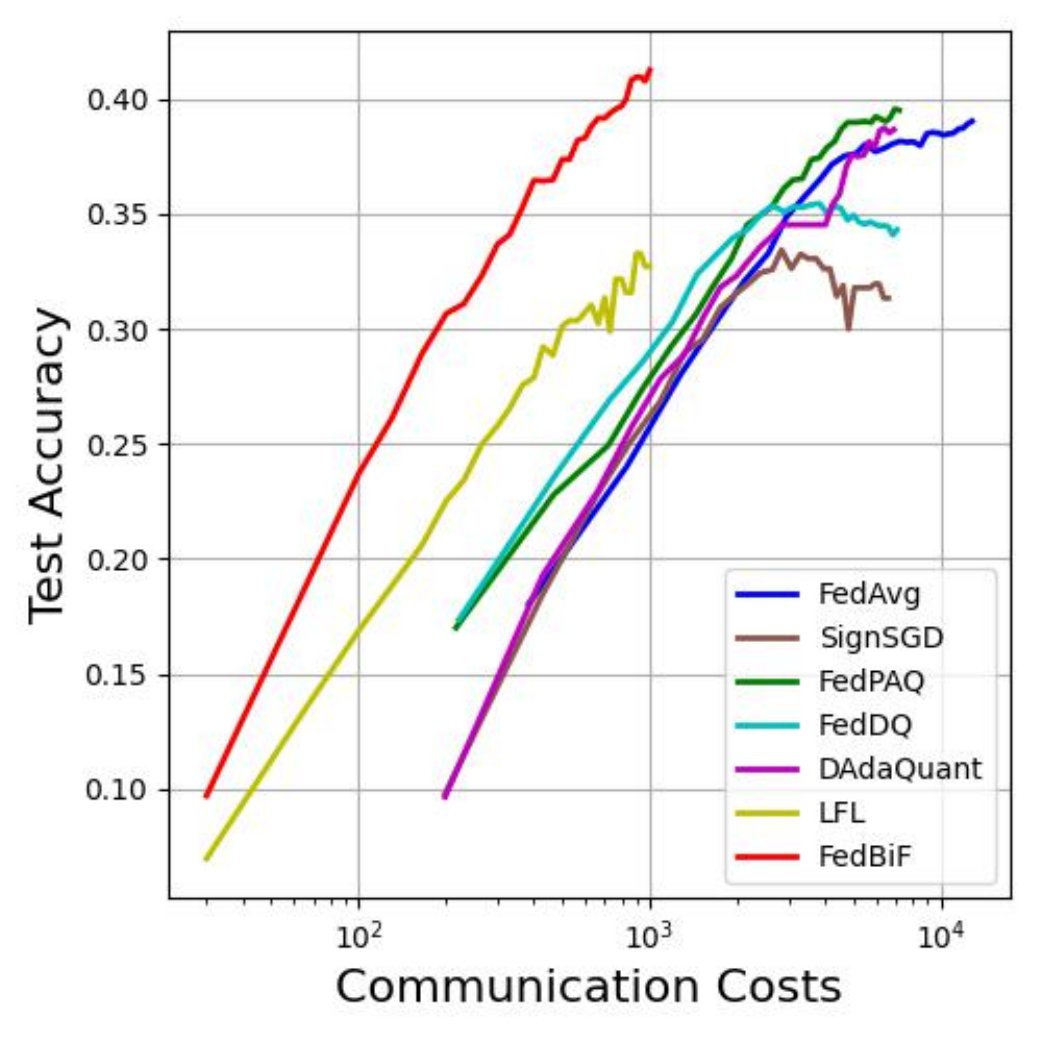}
}
\subfloat[CIFAR-100, Non-IID-1]{
    \includegraphics[scale=0.29]{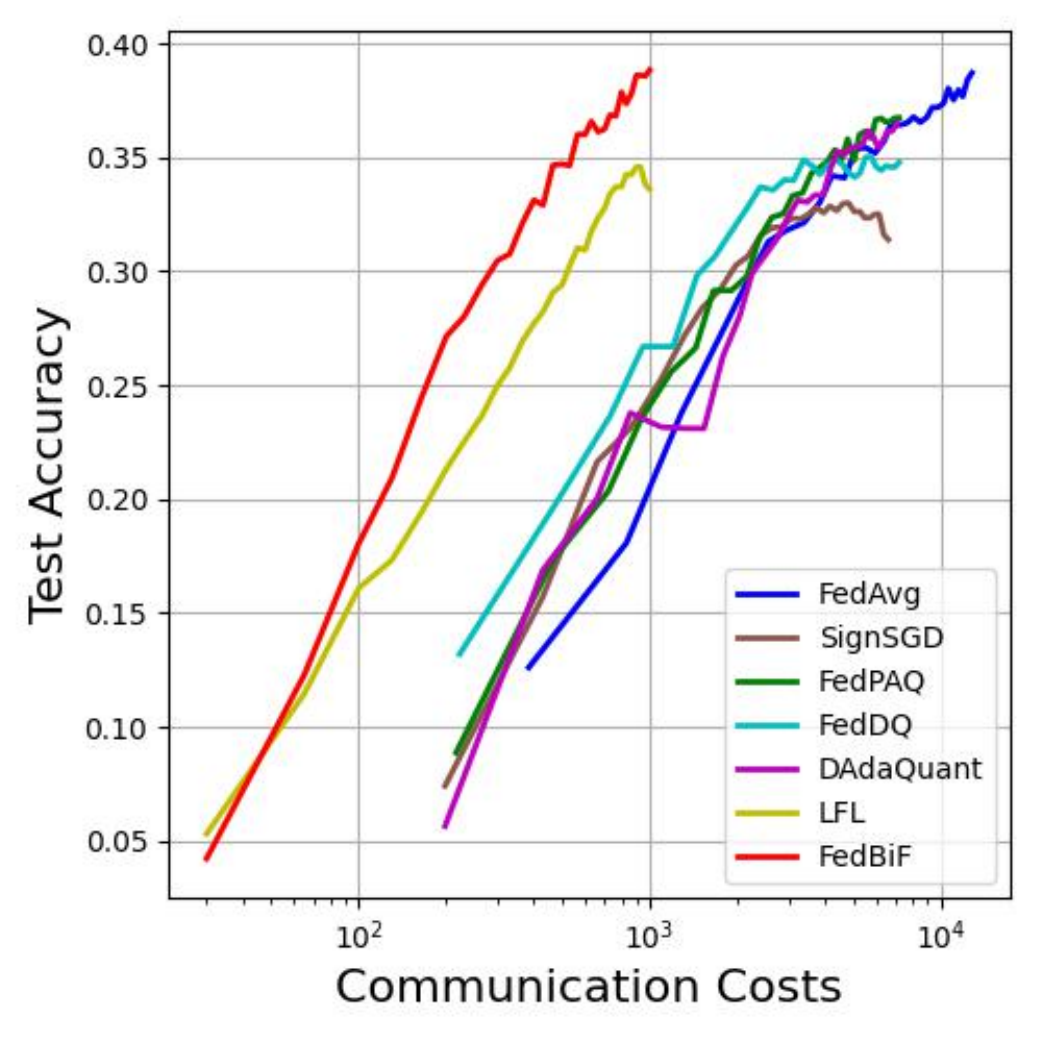}
}
\subfloat[CIFAR-100, Non-IID-2]{
    \includegraphics[scale=0.29]{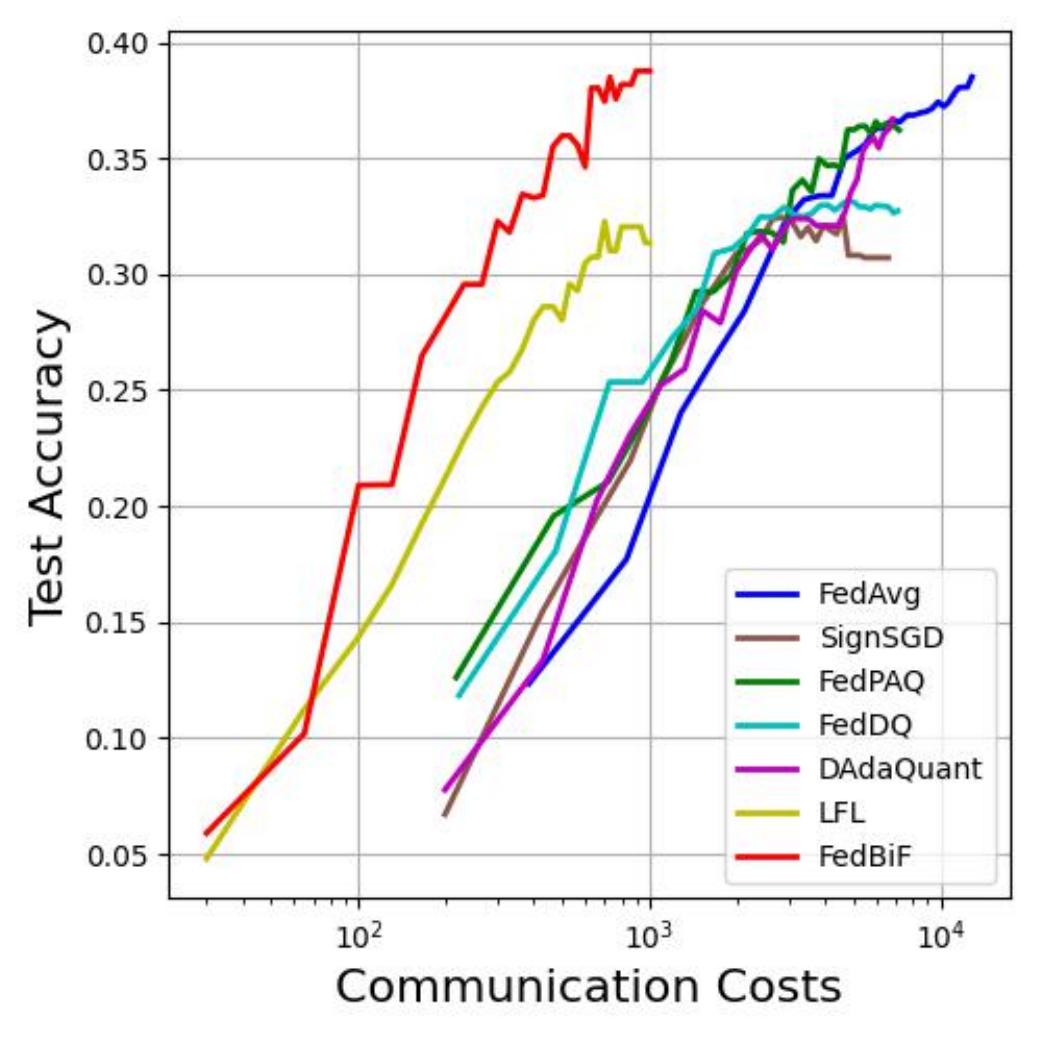}
}
\hfill
\subfloat[TinyImageNet, IID]{
    \includegraphics[scale=0.29]{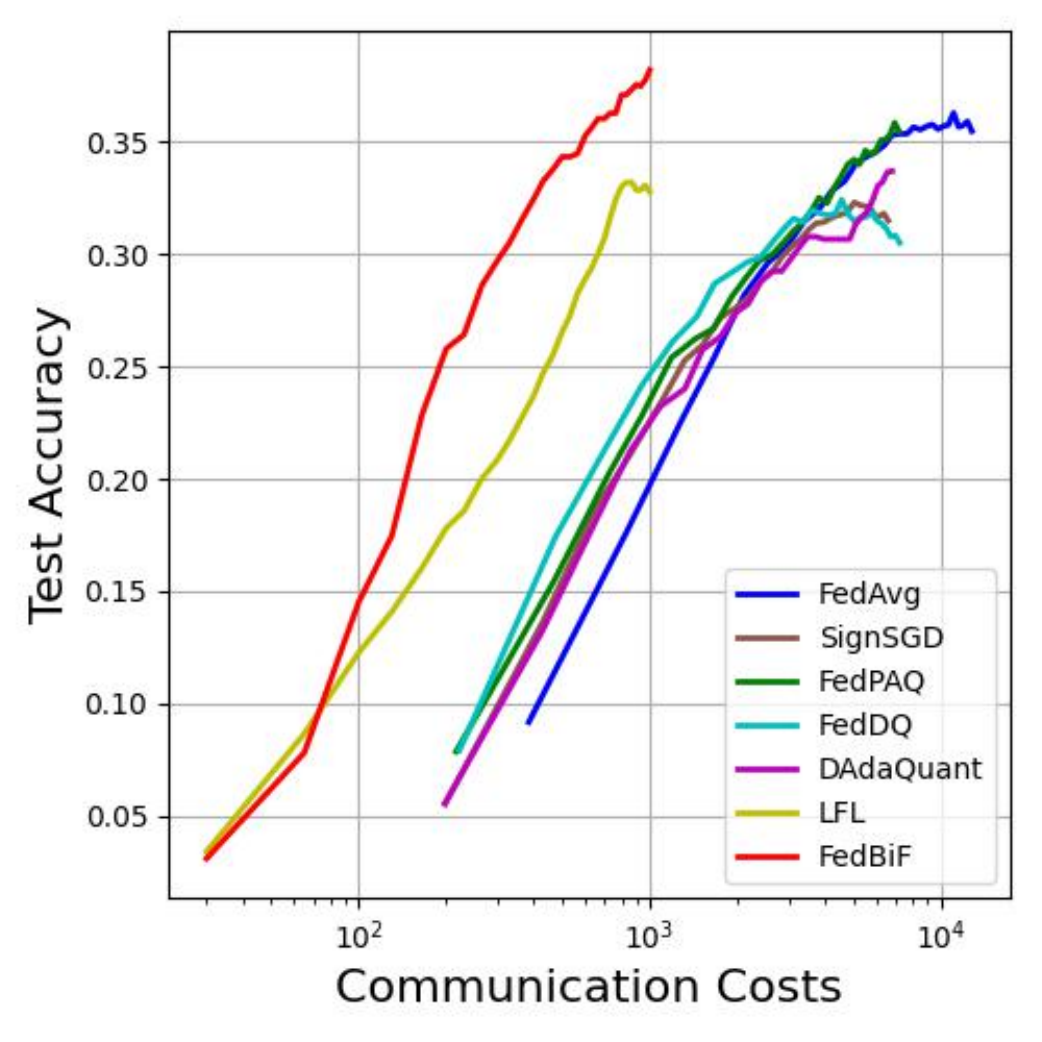}
}
\subfloat[TinyImageNet, Non-IID-1]{
    \includegraphics[scale=0.29]{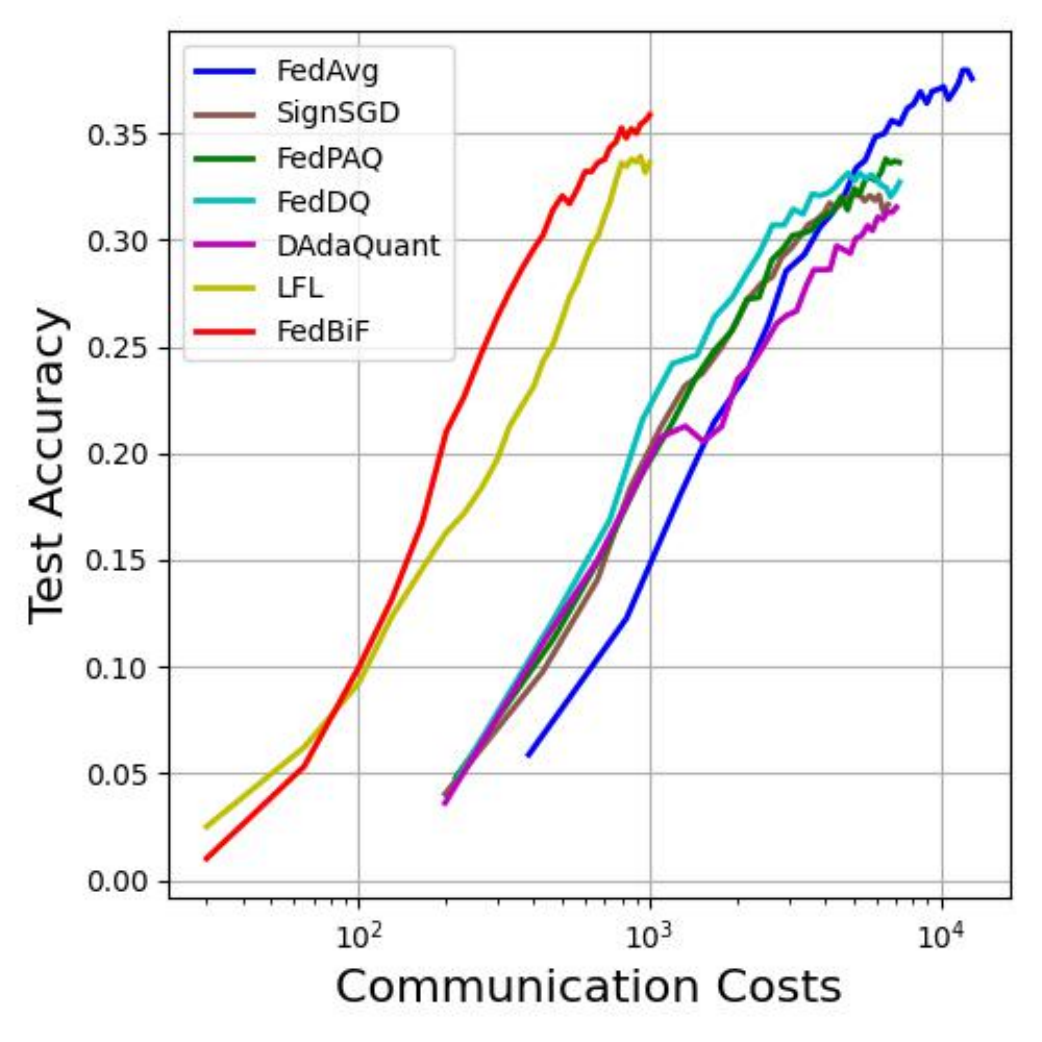}
}
\subfloat[TinyImageNet, Non-IID-2]{
    \includegraphics[scale=0.29]{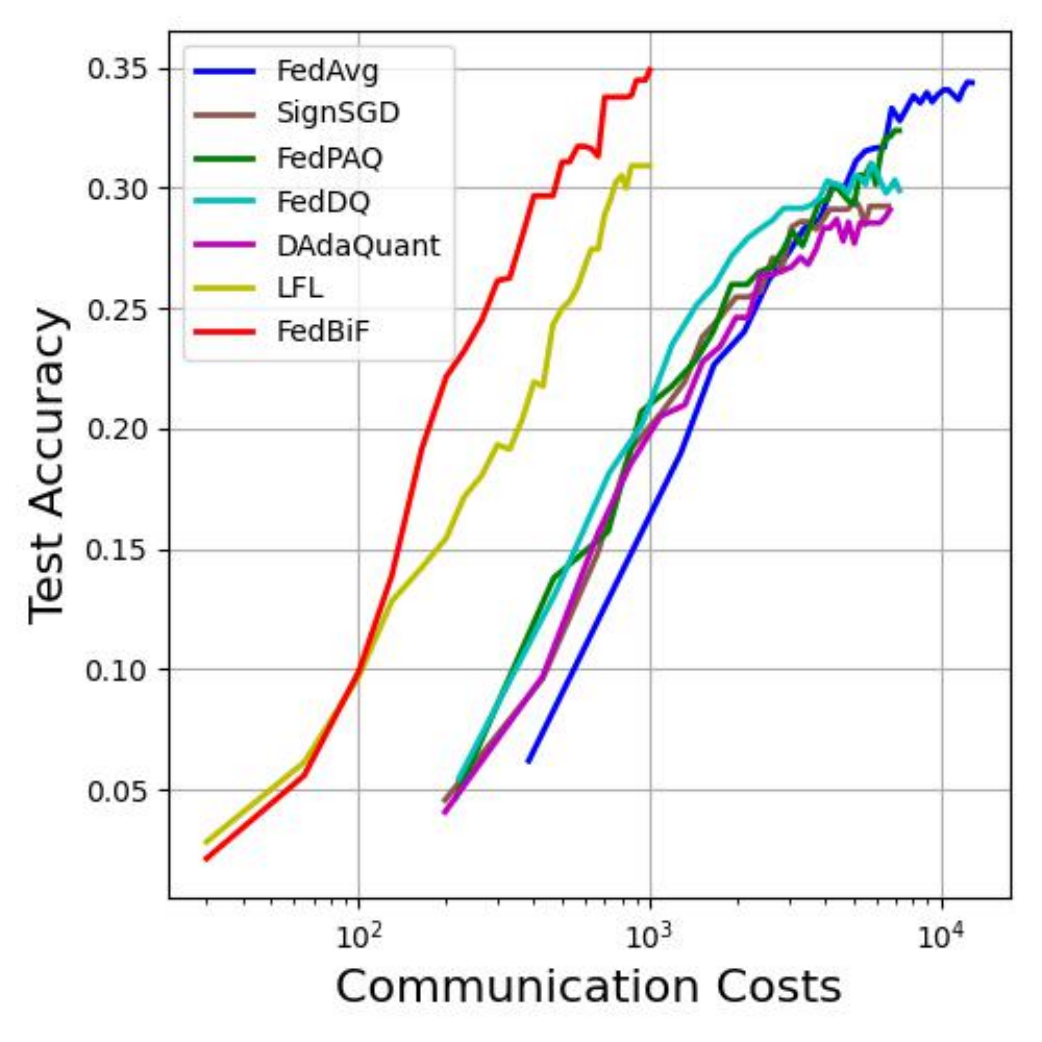}
}
\caption{Convergence curves on CIFAR-10, CIFAR-100, and TinyImageNet under the same communication costs. The x-axis denotes the sum of the bits-per-parameter used in both uplink and downlink communication.}
\label{fig:converge}
\end{figure*}

In this subsection, we compare FedBiF with the baseline methods by three metrics including the test accuracy, the uplink and downlink communication costs. Regardless of the training methods employed, the quantity of parameters to transmit remains constant. Hence, we employ the bit-per-parameter (bpp) as the standard unit for measuring communication costs. All numerical results are reported in Table~\ref{tb:performance}. The convergence curves under the same communication cost are shown in Figure \ref{fig:converge}.

In both IID and Non-IID scenarios, FedBiF achieves comparable accuracy as FedAvg and much better convergence speed than the baselines. As shown in Figure~\ref{fig:converge}, SignSGD achieves 1 bpp uplink communication, while exhibiting slow convergence speed and diminished test accuracy. 
As shown in Table~\ref{tb:performance}, the accuracy of FedPAQ is somewhat lower than that of FedAvg, and the communication cost required is much higher than that of FedBiF.
Both FedDQ and DAdaQuant improve communication efficiency over FedPAQ by adjusting the bit width during training, although the improvements are limited. Moreover, DAdaQuant converges slower due to the lower bit width used in the initial rounds. This issue has also been reported in~\cite{dadaquant}, further illustrating the challenges of post-training quantization with lower bit width. In contrast, FedBiF achieves stronger compression in both uplink and downlink communication without affecting the test accuracy and the convergence speed. Although LFL has the same compression strength, its accuracy is generally lower than FedBiF.

\subsection{Robustness Evaluation}
To comprehensively evaluate the robustness of our proposed method, we conducted additional experiments under two new settings: (1) increasing the number of clients, and (2) adopting an alternative model architecture. The experiment was conducted on CIFAR-10. FedBiF employs a 4-bit quantization and updates only one bit per communication round. LFL uses 1-bit quantization for local model updates and 4-bit quantization for global model updates, resulting in communication overhead equivalent to that of FedBiF. 

\begin{table}[ht]
\centering
\caption{The test accuracy with 200 clients.}\label{tb:client}
\renewcommand{\arraystretch}{1.1}
    
\resizebox{0.5\textwidth}{!}{
\begin{tabular}{lccc}
\toprule[1pt]
& \makebox[0.1\textwidth][c]{IID} & \makebox[0.1\textwidth][c]{Non-IID-1} 
& \makebox[0.1\textwidth][c]{Non-IID-2} 
\\ 
\toprule 
FedAvg & 72.3 ($\pm$0.3) & 64.5 ($\pm$0.2) & 60.8 ($\pm$0.5) \\
LFL    & 68.2 ($\pm$0.4) & 62.3 ($\pm$0.1) & 56.6 ($\pm$0.2) \\
FedBiF & 71.7 ($\pm$0.2) & 64.1 ($\pm$0.3) & 60.1 ($\pm$0.2) \\
\bottomrule[1pt]
\end{tabular}
}
\end{table}

\begin{table}[ht]
\centering
\caption{The test accuracy with DenseNet.}\label{tb:densenet}
\renewcommand{\arraystretch}{1.1}
    
\resizebox{0.5\textwidth}{!}{
\begin{tabular}{lccc}
\toprule[1pt]
& \makebox[0.1\textwidth][c]{IID} & \makebox[0.1\textwidth][c]{Non-IID-1} 
& \makebox[0.1\textwidth][c]{Non-IID-2} 
\\ 
\toprule 
FedAvg & 83.7 ($\pm$0.1) & 77.0 ($\pm$0.1) & 69.0 ($\pm$0.5) \\
LFL    & 80.2 ($\pm$0.1) & 72.0 ($\pm$0.3) & 61.4 ($\pm$0.4) \\
FedBiF & 83.3 ($\pm$0.2) & 77.4 ($\pm$0.2) & 69.5 ($\pm$0.3) \\
\bottomrule[1pt]
\end{tabular}
}
\end{table}

\textbf{Increased Number of Clients.}
To further assess the robustness of our method with respect to the number of clients, we conducted experiments on the CIFAR-10 dataset using 200 clients. Due to the reduced data per client, we increased the number of local training epochs from 3 to 6 to maintain a consistent number of local update steps. All other experimental settings remained unchanged. As shown in Table \ref{tb:client}, increasing the number of clients makes the task more challenging, leading to a decrease in FedAvg's accuracy. In contrast, FedBiF shows no significant accuracy degradation. Under equal communication overhead, FedBiF achieves an average accuracy that is 2\% higher than LFL.

\textbf{Alternative Model Architecture.}
To evaluate the robustness of our method across different model architectures, we also tested it on CIFAR-10 using DenseNet. The hyperparameter settings were kept consistent with the previous experiments. As shown in Table \ref{tb:densenet}, DenseNet achieves higher accuracy than ResNet-18. In this setting, FedBiF still shows no significant accuracy drop compared to FedAvg. Furthermore, with the same communication cost, FedBiF outperforms LFL by an average of 5\%.

\subsection{Ablation Studies}\label{sec:ablation}
In this subsection, we conduct ablation studies to assess the scalability and design of FedBiF. Specifically, our investigation involves varying the quantization bit width, the number of activated bits per round, and the bit selection strategy. In the subsequent discussion, we employ the notation FedBiF-$n$/$m$ to denote a configuration where the bit width is $m$, and $n$ bits are activated in each round. All ablation experiments are performed on the CIFAR-10 dataset.

\textbf{Effect of the Quantization Bit Width.} In FedBiF, the precision of local model parameters is contingent upon the quantization bit width employed in compressing the downlink communication. Table~\ref{tb:bit width} presents the accuracy of FedBiF with various quantization bit widths applied to the CIFAR-10 dataset. It is obvious that the model accuracy rises with the increase of quantization bit width, until a point after which the improvement becomes insignificant. 

\begin{table}[ht]
\centering
\caption{The test accuracy with Varying bit widths.}\label{tb:bit width}
\renewcommand{\arraystretch}{1.1}
    
\resizebox{0.5\textwidth}{!}{
\begin{tabular}{lccc}
\toprule[1pt]
& \makebox[0.1\textwidth][c]{IID} & \makebox[0.1\textwidth][c]{Non-IID-1} 
& \makebox[0.1\textwidth][c]{Non-IID-2} 
\\ 
\toprule 
FedBiF-1/3  & 75.0 ($\pm$ 0.1) & 66.6 ($\pm$ 0.2) & 58.2 ($\pm$ 0.3) \\
FedBiF-1/4  & 77.4 ($\pm$ 0.2) & 69.9 ($\pm$ 0.7) & 64.8 ($\pm$ 0.3) \\
FedBiF-1/5  & 78.1 ($\pm$ 0.2) & 69.9 ($\pm$ 0.2) & 65.8 ($\pm$ 0.2) \\
FedBiF-1/6  & 77.7 ($\pm$ 0.3) & 70.4 ($\pm$ 0.3) & 66.2 ($\pm$ 0.1) \\
\bottomrule[1pt]
\end{tabular}
}
\end{table}

\textbf{Effect of the Number of Activated Bits per Round.}
By default, FedBiF activates one bit during each round. Now, we test the accuracy of FedBiF-2/4 and FedBiF-4/4. In particular, FedBiF-2/4 alternately activates the first two and last two bits, while FedBiF-4/4 activates all four bits. 
As shown in Table~\ref{tb:activate}, in the context of IID scenarios, the number of activated bits has no obvious impact on the final accuracy. Conversely, in the case of Non-IID scenarios, activating more bits per round will actually reduce the test accuracy. This is probably due to the fact that the local model tends to overfit when using more activated bits on heterogeneous local data. Thus, one bit shall be activated in each round for the general FedBiF setting, which is most beneficial in terms of communication cost, computational overhead, and model performance.

\begin{table}[ht]
\centering
\caption{Accuracy of FedBiF with varying number of activated bits.}\label{tb:activate}
\renewcommand{\arraystretch}{1.1}
    
\resizebox{0.5\textwidth}{!}{
\begin{tabular}{lccc}
\toprule[1pt]
& \makebox[0.1\textwidth][c]{IID} & \makebox[0.1\textwidth][c]{Non-IID-1} 
& \makebox[0.1\textwidth][c]{Non-IID-2} 
\\ 
\toprule 
FedBiF-1/4  & 77.4 ($\pm$ 0.2) & 69.9 ($\pm$ 0.7) & 64.8 ($\pm$ 0.3) \\
FedBiF-2/4  & 76.8 ($\pm$ 0.2) & 68.1 ($\pm$ 0.4) & 61.0 ($\pm$ 0.2) \\
FedBiF-4/4  & 77.5 ($\pm$ 0.1) & 66.3 ($\pm$ 0.5) & 59.4 ($\pm$ 0.6) \\
\bottomrule[1pt]
\end{tabular}
}
\end{table}

\textbf{Effect of the Bit Selection Strategy.}
FedBiF employs a simple activated bit selection strategy, iteratively activating the distinct bits across all parameters in different rounds. As shown in Table~\ref{tb:selection}, we report the accuracy of several FedBiF variants. Comparable performance is observed when activated bits are either randomly selected (FedBiF-R1/4) or iteratively cycled through (FedBiF-1/4), highlighting the robustness of FedBiF to bit selection strategies. Furthermore, models trained by iteratively activating different bits significantly outperform those trained on a fixed bit.

\begin{table}[ht]
\centering
\caption{Accuracy of FedBiF with varying bit selection strategies.}\label{tb:selection}
\renewcommand{\arraystretch}{1.1}
    
\resizebox{0.5\textwidth}{!}{
\begin{tabular}{lccc}
\toprule[1pt]
& \makebox[0.1\textwidth][c]{IID} & \makebox[0.1\textwidth][c]{Non-IID-1} 
& \makebox[0.1\textwidth][c]{Non-IID-2} 
\\ 
\toprule 

FedBiF-1/4  & 77.4 ($\pm$ 0.2) & 69.9 ($\pm$ 0.7) & 64.8 ($\pm$ 0.3) \\
FedBiF-R1/4 & 76.6 ($\pm$ 0.1) & 68.8 ($\pm$ 0.6) & 63.2 ($\pm$ 0.5) \\
FedBiF-0001 & 33.2 ($\pm$ 0.1) & 19.9 ($\pm$ 0.5) & 16.8 ($\pm$ 0.4) \\
FedBiF-0010 & 31.3 ($\pm$ 0.2) & 24.5 ($\pm$ 0.5) & 21.0 ($\pm$ 0.6) \\
FedBiF-0100 & 44.5 ($\pm$ 0.4) & 32.0 ($\pm$ 0.7) & 24.6 ($\pm$ 0.2) \\
FedBiF-1000 & 53.8 ($\pm$ 0.6) & 35.4 ($\pm$ 0.8) & 30.2 ($\pm$ 0.4) \\
\bottomrule[1pt]
\end{tabular}
}
\parbox{0.48\textwidth}{\vspace{5pt}
FedBiF-1/4 iteratively activates the 4 bits within each parameter in different rounds. FedBiF-R1/4 randomly activates one of the four bits during each round. {In the remaining variants, FedBiF always trains a specific bit, marked by a position set to 1. For example, FedBiF-1000 always trains the first bit.}
}
\end{table}

\subsection{Model Sparsity}
Conventional gradient descent algorithms struggle to effectively induce sparsity in neural networks, often requiring manual intervention through pruning techniques. In contrast, FedBiF naturally produces sparse models. Specifically, a parameter computed by Eq.~\ref{eq:bit2int} effectively performs automatic pruning when the first bit (i.e., the sign bit) is optimized to one and the remaining bits are optimized to zero. Optimizing a bit to either zero or one is equivalent to adjusting the corresponding virtual bit to a negative or positive value, which is considerably simpler than driving a real-valued parameter exactly to zero.
This mechanism enables FedBiF to eliminate redundant parameters, particularly when the bit width is small. We report the sparsity (i.e., the proportion of zero-valued parameters) of global models trained by FedBiF with a quantization bit width of 2. As shown in Table~\ref{tb:sparsity}, FedBiF achieves higher sparsity on the FMNIST and SVHN datasets, consistent with the observation that simpler datasets tend to contain more redundant parameters. Additionally, the resulting model sparsity can be exploited to further reduce downlink communication costs. In contrast, the models trained by baseline methods exhibit zero sparsity.

\begin{table}[h]
  \centering
  \caption{The average sparsity of models trained by FedBiF.}\label{tb:sparsity}
  \scalebox{1.0}{
  \begin{tabular}{cccc}
    \toprule
     & \makebox[0.1\textwidth][c]{IID} & \makebox[0.1\textwidth][c]{Non-IID-1} & \makebox[0.1\textwidth][c]{Non-IID-2} \\
    \midrule
    FMNIST      & 47\% & 56\%  & 49\% \\
    SVHN        & 58\% & 54\%  & 62\% \\
    CIFAR-10    & 32\% & 37\%  & 39\% \\
    CIFAR-100   & 26\% & 22\%  & 24\% \\
    TinyImageNet   & 22\% & 18\%  & 20\% \\
    
    \bottomrule
  \end{tabular}}
\end{table}

\subsection{Memory and Computational Overhead Analysis}
First, we analyze the memory overhead introduced by the local model in FedBiF. Importantly, FedBiF activates only a single bit per round; only this activated bit requires training via a virtual bit for updates, while the frozen bits can be pre-computed and stored, thereby reducing both memory and computational costs. Consequently, FedBiF stores one virtual bit (a floating-point number) and the pre-computed contribution of the frozen bits (an $m$-bit integer) for each weight. Moreover, since the binary value of the virtual bit is determined solely by its sign, it does not require high precision. Therefore, we argue that representing the virtual bit with lower-precision formats, such as FP16 or FP8, is sufficient to preserve numerical accuracy while further reducing local memory usage.

Second, we analyze the computational overhead of the local model in FedBiF. As discussed above, FedBiF activates only one bit per round, while the frozen bits can be pre-computed to reduce additional computation. This pre-computation is performed once before local training begins, and its computation is negligible compared to that of the training stage. To reconstruct each weight, an additional summation is required, i.e., $\theta = s + \alpha_i \times b$, where $s$ denotes the pre-computed result of all frozen bits, $\alpha_i = \alpha \times 2^i$ represents the scaling factor of the activated bit, and $b$ is its binary value. Importantly, the cost of this reconstruction is independent of batch size (BS), whereas the computational cost of the model itself grows with batch size. In standard local training, relatively large batch sizes (e.g., 64) are typically employed. As shown in Table \ref{tb:mac}, under this setting (BS=64), the relative increase in computational cost introduced by FedBiF is only 0.03\%, which is effectively negligible.  
In Table \ref{tb:mac}, we report computational overhead in terms of multiply–accumulate operations (MACs), the standard metric for measuring neural network computation. Each MAC corresponds to one multiplication followed by one addition.

\begin{table}[h]
  \centering
  \caption{MACs of ResNet18 on CIFAR-10 under various Batch Sizes (BS).}\label{tb:mac}
  \scalebox{1.0}{
  \begin{tabular}{ccccc}
    \toprule
     & \makebox[0.06\textwidth][c]{BS=1} & \makebox[0.06\textwidth][c]{BS=4} & \makebox[0.06\textwidth][c]{BS=16} & \makebox[0.06\textwidth][c]{BS=64} \\ \midrule
FedAvg & 558M                   & 2232M                  & 8926M                   & 35704M                  \\ 
FedBiF & 569M                   & 2243M                  & 8937M                   & 35715M                  \\ \midrule
Relative Increase  & +2.00\%                & +0.50\%                & +0.13\%                 & +0.03\%           \\
\bottomrule
  \end{tabular}}
\end{table}

\section{Conclusion and Future Work}\label{sec:conclusion}
In this paper, we address the limitations of post-training quantization in communication-efficient federated learning (CEFL). Specifically, we propose Federated Bits Freezing (FedBiF), a novel framework that enables bit-by-bit optimization of quantized parameters during local training. By selectively updating a single bit per parameter while freezing the rest, FedBiF achieves high-precision weight representation at a low communication cost.
Our extensive experiments on multiple benchmark datasets demonstrate that FedBiF consistently outperforms existing CEFL methods in both uplink and downlink compression without compromising model accuracy or convergence. Furthermore, FedBiF naturally induces sparsity in the learned models, which can potentially enhance generalization.
Overall, FedBiF offers a promising direction for better CEFL methods. In addition, we notice that there are still some designs that can be further improved for FedBiF. For example, we believe that the bit selection strategy can be designed in an adaptive fashion to achieve better convergence speed. Also, the quantization step size can be designed to be learnable, which may further improve model accuracy and speed up convergence. We leave these for future work.

\section*{Acknowledgements}
This work is supported by the National Key Research and Development Program of China under grant 2024YFC3307900; the National Natural Science Foundation of China under grants 62376103, 62302184, 62436003 and 62206102; Major Science and Technology Project of Hubei Province under grant 2024BAA00;Hubei Science and Technology Talent Service Project under grant 2024DJC07;
and Ant Group through CCF-Ant Research Fund.

\bibliographystyle{plain}
\bibliography{main}

\begin{IEEEbiography}[{\includegraphics[width=1in,height=1.25in,clip,keepaspectratio]{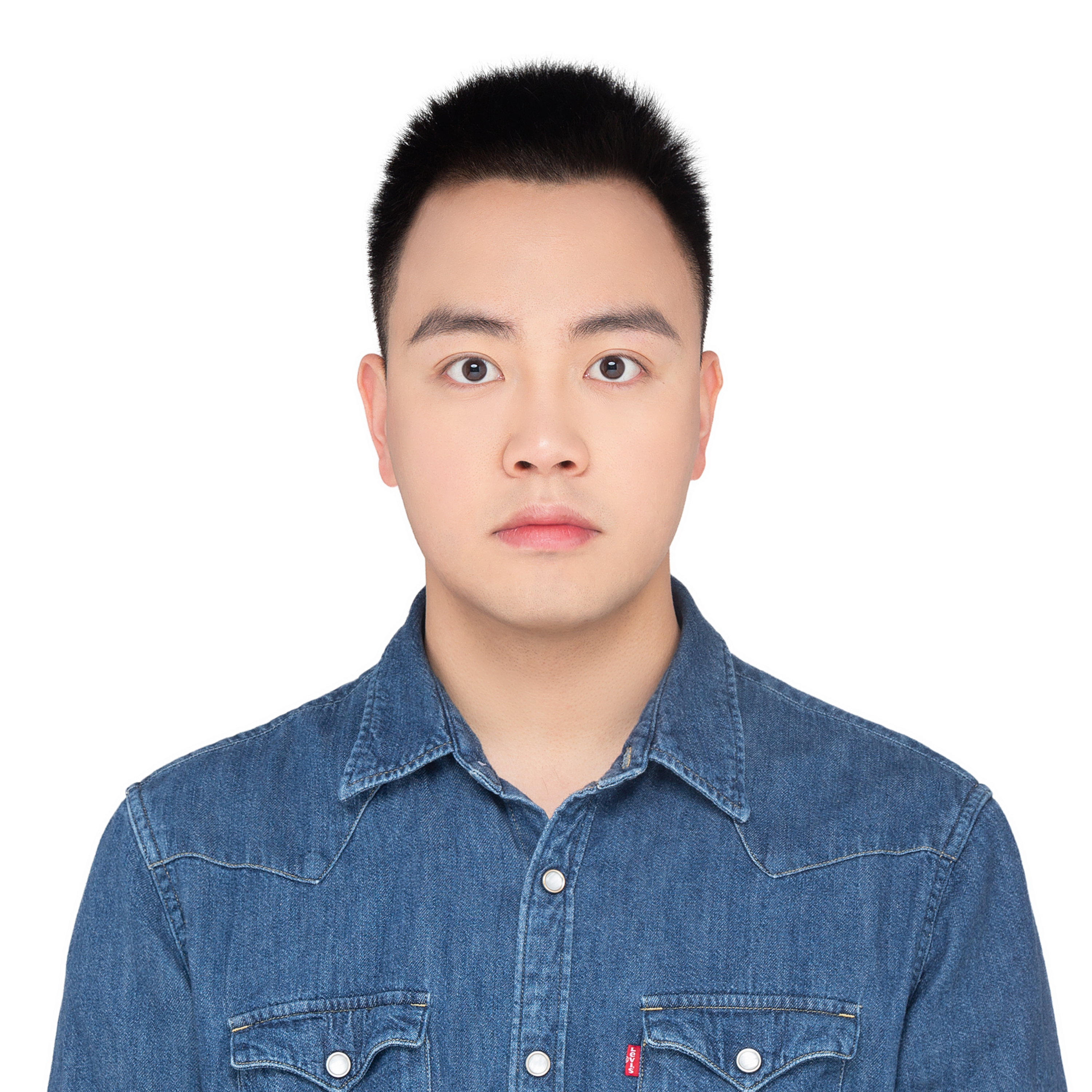}}]{Shiwei Li} received the bachelor’s degree from Beijing University of Posts and Telecommunications, China, in 2021. He is currently a PhD student at Huazhong University of Science and Technology. His research interests include model compression, federated learning, recommender systems, and large language models.\end{IEEEbiography}

\begin{IEEEbiography}[{\includegraphics[width=1in,height=1.25in,clip,keepaspectratio]{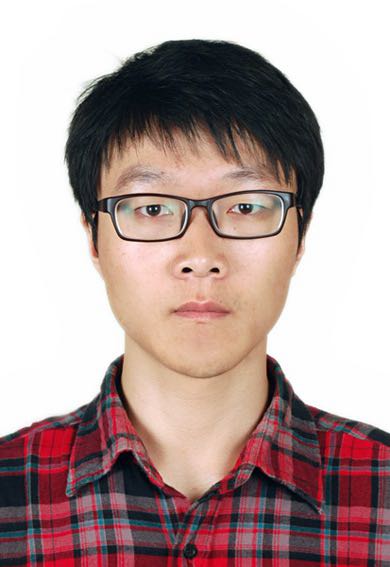}}]{Qunwei Li} received the Ph.D. degree in electrical and computer engineering from Syracuse University, Syracuse, NY, USA, in 2018. He was a postdoctoral researcher in the Center for Applied Scientific Computing (CASC) at Lawrence Livermore National Laboratory (LLNL), CA, USA, from 2018 to 2019. He worked at Ant Group, China, on machine learning algorithms for recommender systems from 2019 to 2025. His research interests include decision making, deep learning, and optimization algorithms. Dr. Li received the Syracuse University Graduate Fellowship Award in 2014 and the All University Doctoral Prize in 2018 by Syracuse University for superior achievement in completed dissertations.\end{IEEEbiography}

\begin{IEEEbiography}[{\includegraphics[width=1in,height=1.25in,clip,keepaspectratio]{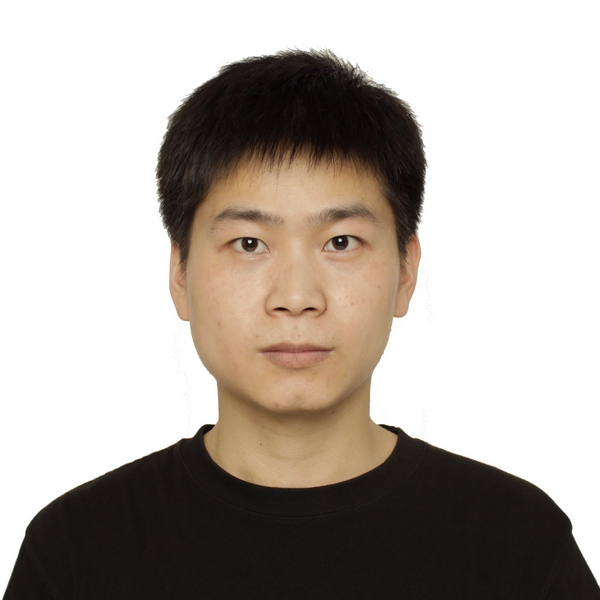}}]{Haozhao Wang} (Member, IEEE) received the bachelor’s degree from the University of Electronic Science and Technology, and the PhD degree from
 the School of Computer Science and Technology, Huazhong University of Science and Technology, in 2021. He is an assistant professor at the School of Computer Science and Technology, Huazhong University of Science and Technology. He worked as a research assistant with the Department of Computing, The Hong Kong Polytechnic University from 2018 to 2021. His research interests include distributed machine learning, federated learning, and AI security.\end{IEEEbiography}

\begin{IEEEbiography}[{\includegraphics[width=1in,height=1.25in,clip,keepaspectratio]{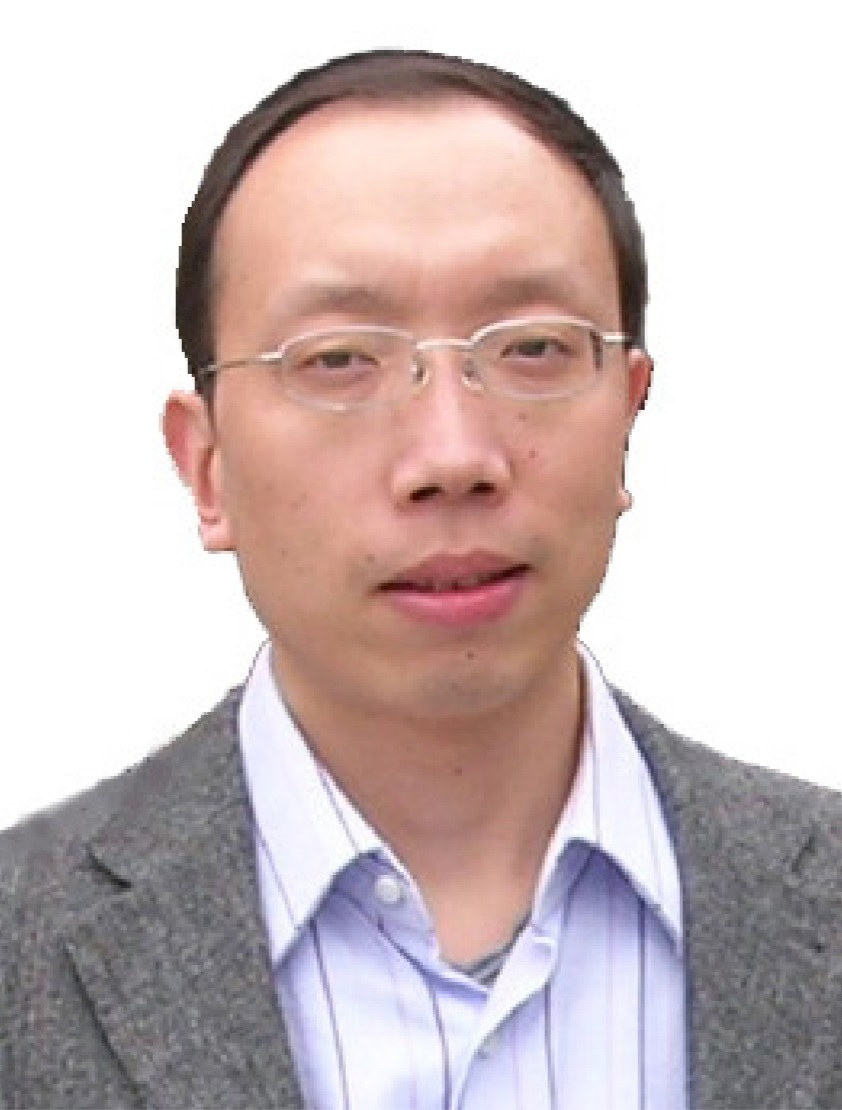}}]{Ruixuan Li} (Member, IEEE) received the BS, MS, and PhD degrees in computer science from the Huazhong University of Science and Technology, China, in 1997, 2000, and 2004, respectively. He is a professor with the School of Computer Science and Technology, Huazhong University of Science and Technology. He was a visiting researcher with the Department of Electrical and Computer Engineering, University of Toronto from 2009 to 2010.His research interests include cloud and edge computing, Big Data
 management, and distributed system security. He is a member of ACM.\end{IEEEbiography}

\begin{IEEEbiography}[{\includegraphics[width=1in,height=1.25in,clip,keepaspectratio]{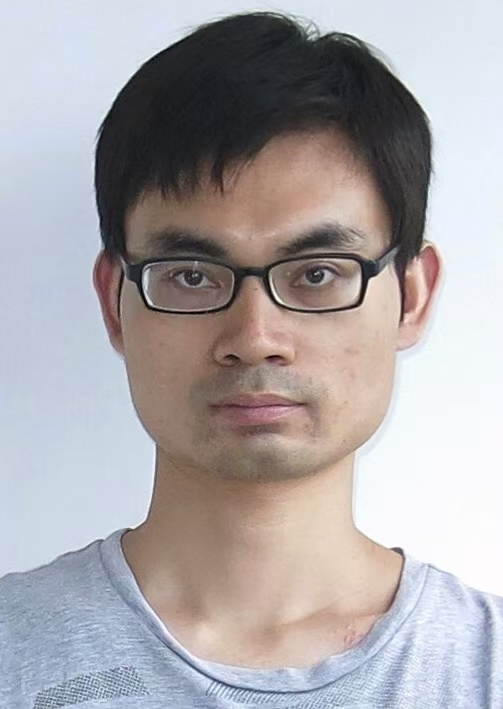}}]{Jianbin Lin} received his Master's degree in Computer Science from Beijing Normal University, China, in 2013. Since 2017, he has been serving as a Machine Learning Engineer at Ant Group, China, where he focuses on developing machine learning algorithms for recommender systems. His research interests focus on addressing challenges in industrial-scale recommendation systems and distributed machine learning implementations.\end{IEEEbiography}

\begin{IEEEbiography}[{\includegraphics[width=1in,height=1.25in,clip,keepaspectratio]{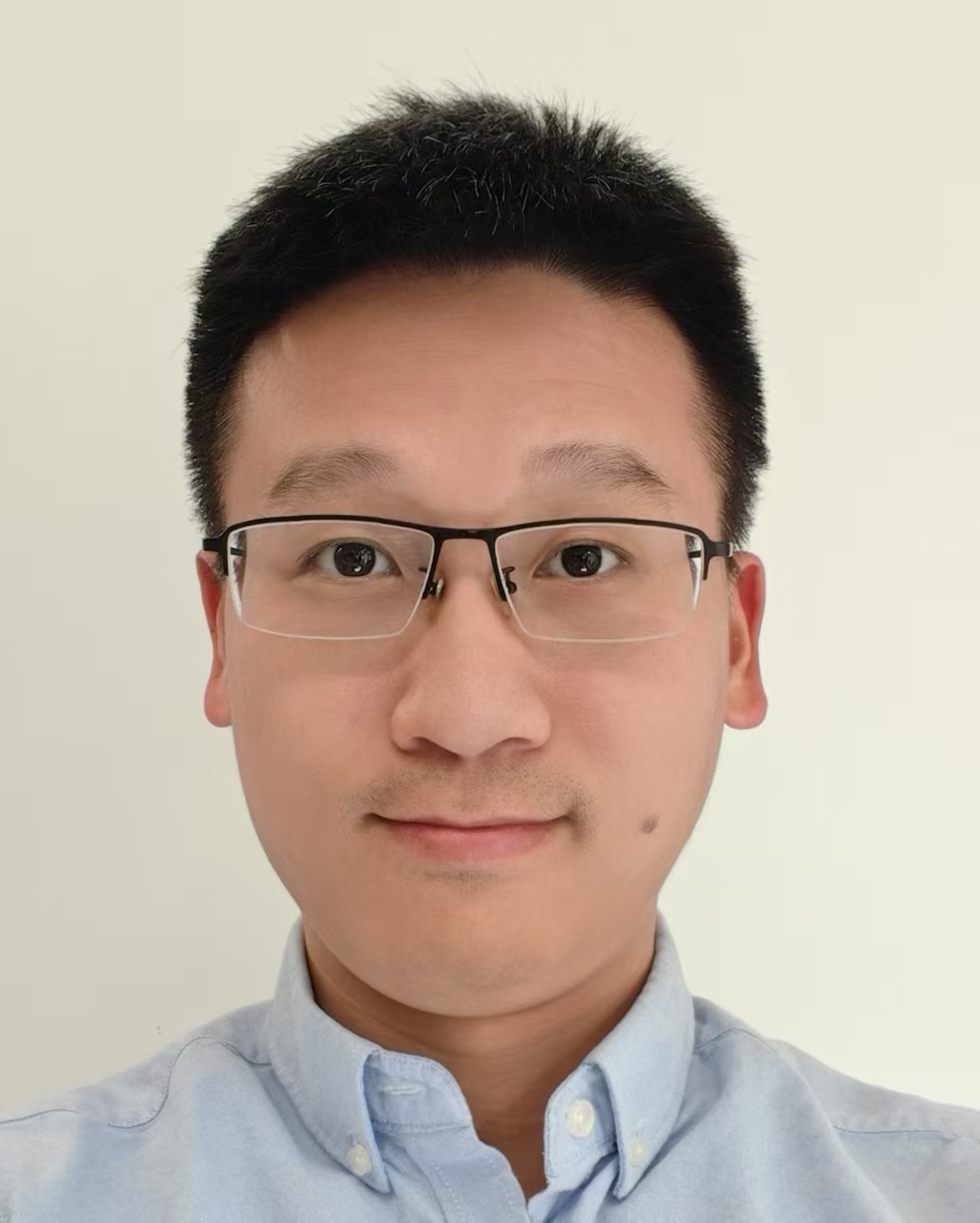}}]{Wenliang Zhong} received his Ph.D. degree in computer science from the Hong Kong University of Science and Technology. Following his graduation, he joined Alibaba Group and Ant Group successively, and currently serves as the Director of the Recommendation Algorithms Department at Alipay. His research focuses encompass large-scale optimization algorithms, recommender systems, and LLMs, with over 10 years of experience applying cutting-edge academic solutions to industrial practice.\end{IEEEbiography}
\end{document}